\documentclass[12pt]{article}

\usepackage{newtxtext,newtxmath}

\usepackage{graphicx}

\usepackage[letterpaper,margin=1in]{geometry}

\renewenvironment{abstract}
	{\quotation}
	{\endquotation}

\date{}

\makeatletter
\renewcommand{\fnum@figure}{\textbf{Figure \thefigure}}
\renewcommand{\fnum@table}{\textbf{Table \thetable}}
\makeatother

\usepackage{scicite}

\usepackage{url}

\def\scititle{Towards bio-inspired visual navigation: a neuromorphic navigation paradigm for MAVs}
\title{\bfseries \boldmath \scititle}

\author{
	Junjie Jiang$^{1}$,
	Ankang Zhou$^{2}$,
	Zheng Fang$^{1,2\ast}$\and
	\small$^{1}$Faculty of Robot Science and Engineering, Northeastern University, Shenyang 110819, China.\and
	\small$^{2}$ College of Information Science and Engineering, Northeastern University, Shenyang 110819, China.\and
	\small$^\ast$Corresponding author. Email: fangzheng@mail.neu.edu.cn\and
}

\begin{document} 

\maketitle

\begin{abstract} \bfseries \boldmath
Biological visual navigation achieves remarkable robustness at extremely low power, whereas frame-based robotic systems relying on artificial neural networks remain limited in both respects, particularly for energy-constrained micro aerial vehicles (MAVs) operating in complex environments.
Inspired by the event-driven, sparse, and asynchronous principles of biological visual systems, neuromorphic sensing and computation provide a promising alternative for robust and energy-efficient navigation.
In this paper, we propose the first end-to-end event-driven neuromorphic paradigm for MAV visual navigation in complex environments by integrating neuromorphic vision, spiking neural networks, and neuromorphic processors. 
The proposed method first extracts the environmental representation from event streams using a spiking perceptual representation module. 
This representation is fed into a spiking navigation module, where guidance learning and population coding enable efficient policy learning with few time steps. 
The navigation module then outputs a waypoint converted into executable trajectory points.
Experiments show that the proposed neuromorphic method achieves navigation performance comparable to ANN-based methods while markedly reducing power consumption. 
The spiking perceptual module consumes only 19.57\% of the power of its ANN counterpart, while the spiking navigation module deployed on Intel Loihi consumes only 12.37\% of that required by its ANN counterpart running on NVIDIA Orin NX.
Beyond prior neuromorphic systems focused mainly on relatively simple navigation subtasks, we demonstrate a complete neuromorphic visual navigation system spanning simulation and real-world deployment.
Our results establish the feasibility of the neuromorphic paradigm for autonomous navigation and provide a bio-inspired pathway toward robust and low-power MAV navigation.
\end{abstract}

\subsection*{Introduction}
\begin{figure} 
	\centering
	\includegraphics[width=\textwidth]{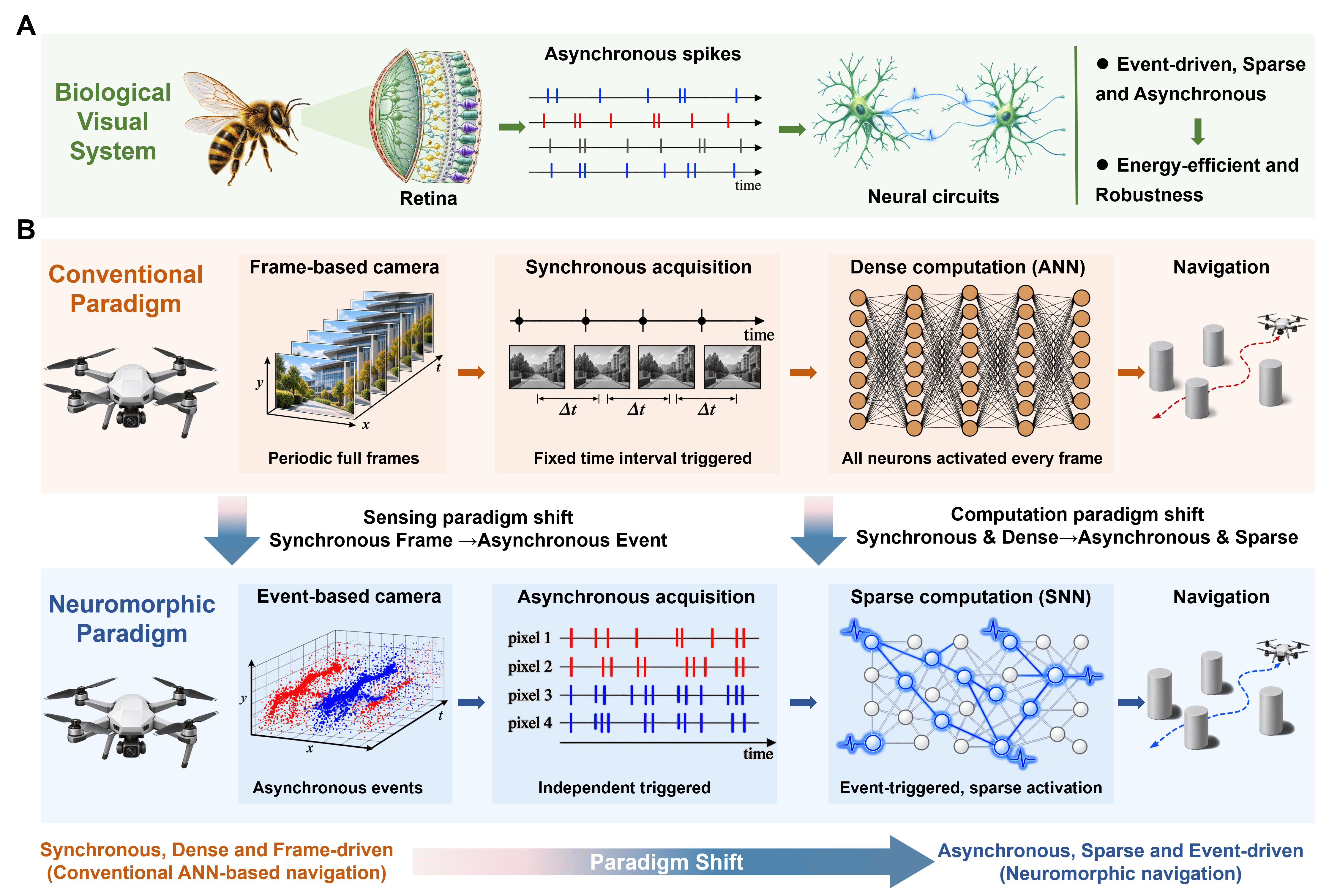}
	\caption{\textbf{Visual navigation mechanisms across biological, conventional, and neuromorphic paradigms.} (\textbf{A}) The biological visual navigation paradigm relies on the retina and neural circuits. Visual information is sensed asynchronously at the pixel level and processed through sparse, event-driven neural activity to produce navigation behaviors. (\textbf{B}) Comparison between conventional and neuromorphic visual navigation paradigms for MAVs. The conventional visual navigation paradigm relies on frame-based cameras and artificial neural networks. Visual information is acquired synchronously as image frames and processed through dense, clock-driven computation to generate navigation actions. The neuromorphic visual navigation paradigm employs event-based vision sensors and spiking neural networks. Inspired by biological visual processing, it captures visual information asynchronously at the pixel level and processes it through sparse, event-driven computation to generate navigation actions.}
	\label{fig:1} 
\end{figure}

In nature, insects and birds can rely on compact visual systems and efficient neural information processing to achieve long-duration, robust navigation in complex environments under extremely limited computational and energy resources~\cite{collett2002memory, bhagavatula2011optic}.
This ability to perform rapid perception, real-time decision-making, and stable flight under severe resource constraints highlights a significant gap between the navigation capabilities of current small aerial robots and those of flying animals.
Over the past decade, advances in visual sensors and artificial neural networks have substantially improved the capabilities of flying robots, especially small rotorcraft MAVs, in environmental perception~\cite{qin2018vins,teed2023deep}, autonomous obstacle avoidance~\cite{song2023learning,loquercio2021learning,zhou2022swarm}, and closed-loop control~\cite{song2022policy,o2022neural}.
However, existing MAVs still fall short of the robustness and energy efficiency exhibited by biological visual navigation systems in complex scenarios.
Current mainstream navigation systems based on conventional frame-based vision and artificial neural networks~\cite{kaufmann2023champion,yu2025spatial,yu2025depth} acquire image frames at fixed intervals and process visual information through dense and synchronous computation, as shown in Fig.~\ref{fig:1}B.
Many existing systems~\cite{kaufmann2023champion, wu2026precise} rely on visually salient ground features, such as fiducial markers or high-contrast objects, to maintain reliable perception during highly agile flight. 
In contrast to the sparse and asynchronous sensing and computation of biological systems, as illustrated in Fig.~\ref{fig:1}A, this frame-driven paradigm makes it difficult to simultaneously achieve robust perception and high energy efficiency on high-speed, resource-constrained MAVs. 
More fundamentally, this mismatch in sensing and computation principles may not only constrain navigation performance in increasingly challenging environments, but also limit the potential of conventional architectures to evolve toward more adaptive, efficient, and biologically inspired autonomy.

Neuromorphic technologies~\cite{chakravarthi2024recent,gallego2020event,roy2019towards} offer a promising route to narrowing the gap between robotic navigation systems and biological navigation systems.
Unlike conventional frame-based vision sensors and artificial neural networks, neuromorphic vision sensors, known as event cameras, and spiking neural networks are characterized by event-driven operation, sparse activation, and asynchronous processing, which are intrinsically aligned with biological visual perception and neural computation.
Event cameras generate events only when pixel-level brightness changes occur, enabling visual information to be captured with microsecond-level temporal resolution while substantially reducing redundant data.
In contrast, spiking neural networks transmit and process information through asynchronous discrete spikes, and can significantly reduce computational power consumption, especially when deployed on neuromorphic processors.
In recent years, neuromorphic technologies have attracted increasing attention and achieved significant progress~\cite{klenk2024deep,guan2025deio,zhou2021event,niu2025esvo2,zhu2025depth,ghosh2025event,hu2025ede,gehrig2024low,li2022asynchronous,gehrig2023recurrent,yen2024tracking,fang2021deep,fang2023parallel,yao2025scaling,luo2024integer}.
Event cameras have demonstrated advantages over conventional frame-based cameras in challenging scenarios involving high-speed motion and complex illumination, and have been widely applied to visual tasks such as pose estimation~\cite{klenk2024deep,guan2025deio,zhou2021event,niu2025esvo2}, depth estimation~\cite{zhu2025depth,ghosh2025event,hu2025ede}, and object detection~\cite{gehrig2024low,li2022asynchronous,gehrig2023recurrent,yen2024tracking}.
In parallel, spiking neural networks have shown the potential to achieve performance comparable to artificial neural networks across various tasks~\cite{fang2021deep,fang2023parallel,yao2025scaling,luo2024integer}, while substantially reducing energy consumption.
In particular, the rapid development of dedicated neuromorphic processors~\cite{davies2018loihi, akopyan2015truenorth, painkras2013spinnaker,pedersen2019spinnaker,richter2023speck} for spike-based computation has made sparse asynchronous computing and practical deployment of spiking neural networks increasingly feasible. 
These advances highlight the promise of neuromorphic technologies for efficient and robust visual navigation.

However, existing neuromorphic navigation studies~\cite{vemprala2021representation, bhattacharya2024monocular,jesusanmi2024investigating,jiang2023neuro} often introduce neuromorphic components only at either the sensing or computing stage, \textit{and have not yet established a unified end-to-end neuromorphic navigation framework}.
Some studies~\cite{vemprala2021representation, bhattacharya2024monocular} use event cameras to acquire visual information, but still rely on artificial neural networks for subsequent processing. 
Others~\cite{jesusanmi2024investigating,jiang2023neuro} adopt spiking neural networks for decision-making and inference, while their visual inputs still come from conventional frame-based images.
This mismatch between sensing and computation weakens the core advantages of neuromorphic systems, including sparsity, asynchrony, and low-redundancy processing, and also limits their potential for low-latency and energy-efficient navigation.
In contrast, the integration of event cameras and spiking neural networks can preserve consistent properties across sensing and computation, providing a more promising technical route toward bio-inspired and energy-efficient autonomous visual navigation.
With the development of neuromorphic processors, researchers have begun to explore neuromorphic robotic systems and have made preliminary progress in perception and control tasks~\cite{vitale2021event,salt2019parameter,paredes2024fully,jiang2025fully}.
However, due to the still immature hardware and software ecosystem, the lack of structural information in event streams, the limitations of spiking neural networks in efficient representation and end-to-end training, and the limited number of neurons and operators supported by neuromorphic processors, existing studies have largely remained at the level of relatively simple navigation subtasks, for example pose estimation or motion control, etc.
A neuromorphic navigation paradigm for MAVs that integrates event-based sensing, spike-based computation, and neuromorphic hardware deployment in real-world complex environments remains an open challenge.

In this paper, we present an end-to-end event-driven neuromorphic paradigm for visual navigation of MAVs in complex environments for the first time.
By integrating an event camera, a spiking neural network, and a neuromorphic processor into a closed-loop navigation system, the proposed paradigm mimics biological visual sensing and computation as shown in Fig.~\ref{fig:1}B, providing a system-level pathway toward robust, low-power, and real-time visual navigation for MAVs. 
The main contributions of this paper are as follows:

First, we present the first end-to-end event-driven neuromorphic visual navigation paradigm for MAVs that has been successfully transferred from simulation to real-world deployment. 
By integrating neuromorphic sensing, spike-based computation, and neuromorphic hardware deployment, the proposed framework demonstrates the practical feasibility of end-to-end neuromorphic navigation on resource-constrained MAV platforms.

Second, we develop an onboard spiking perceptual representation module that converts asynchronous event streams into compact geometric features for MAV navigation. 
By compensating for the absence of explicit environmental geometric structure in raw event streams, this module provides navigation-relevant structural representations while alleviating the sim-to-real gap induced by rendering discrepancies between simulation and real-world scenes.

Third, we propose an end-to-end MAV navigation policy based on spiking neural networks, enabled by efficient guidance learning and population coding. 
Guidance learning improves policy training efficiency, while population coding supports informative spike-based representations with a small number of time steps. 
The learned policy is further deployed on a neuromorphic processor, establishing an energy-efficient and real-time navigation system.

\subsection*{Results}
\begin{figure} 
	\centering
	\includegraphics[width=0.95\textwidth]{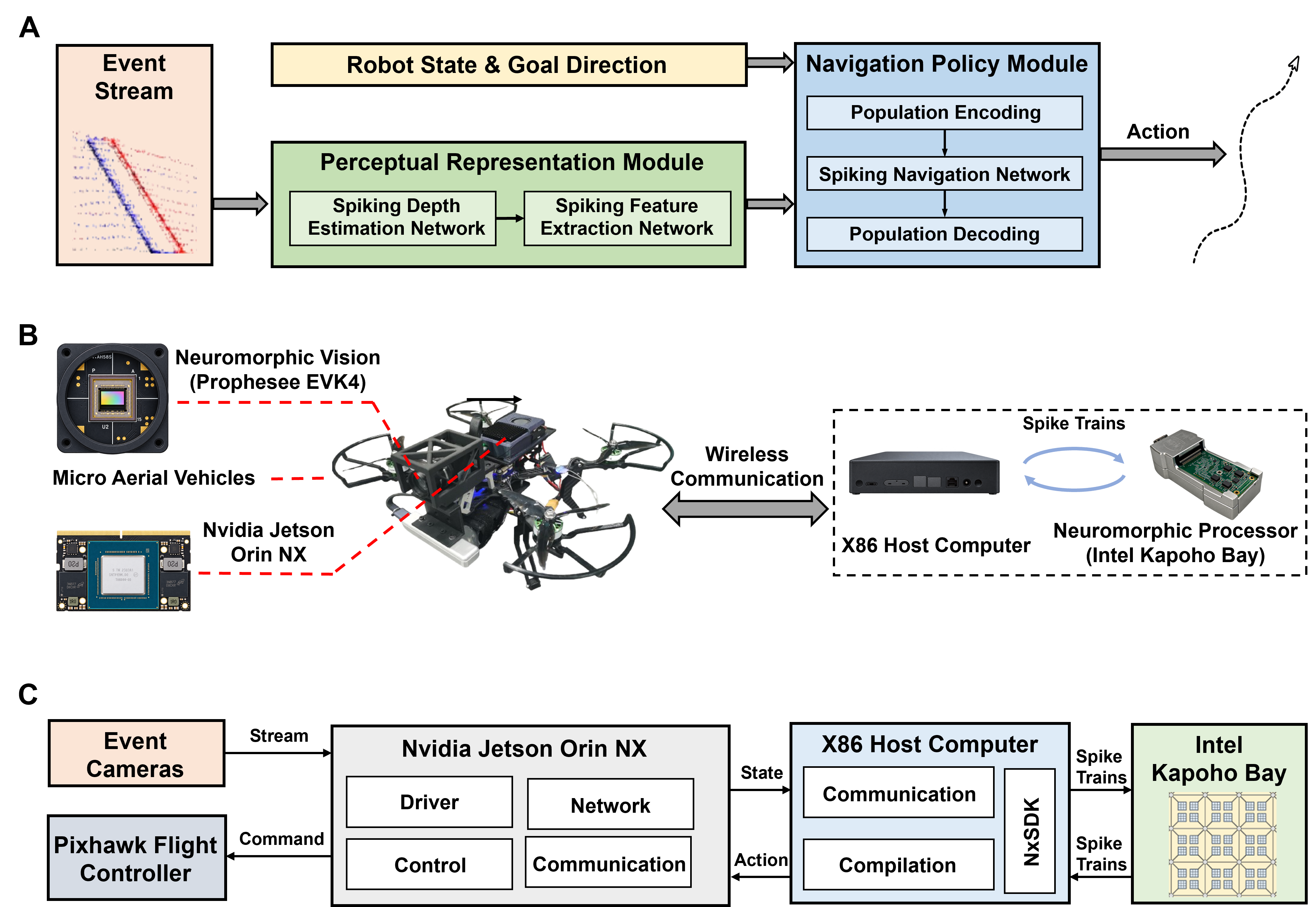}
	\caption{\textbf{End-to-end event-driven neuromorphic MAV navigation system.} (\textbf{A}) Architecture of the proposed end-to-end event-driven neuromorphic navigation system. The system comprises a spiking perceptual representation module and a spiking navigation policy module. The perceptual module converts event streams into compact, geometry-aware features, while the navigation policy module maps these visual features, together with the robot state and goal direction, to navigation actions. (\textbf{B}) Key hardware components of the neuromorphic system. The platform consists of a neuromorphic vision sensor (Prophesee EVK4), an NVIDIA Jetson Orin NX, an x86 host computer, a neuromorphic processor (Intel Kapoho Bay), and a micro aerial vehicle. The NVIDIA Jetson Orin NX communicates wirelessly with the x86 host computer. (\textbf{C}) Deployment and communication pipeline of the proposed system. Owing to the limited neuron models supported by Intel Kapoho Bay, the spiking perceptual representation module is deployed on the NVIDIA Jetson Orin NX, whereas the spiking navigation policy module is deployed on Intel Kapoho Bay. The Jetson Orin NX acquires the event stream and transmits control commands to the Pixhawk flight controller. The Jetson Orin NX and the x86 host computer exchange system information wirelessly. The x86 host computer exchanges spike trains with Intel Kapoho Bay through NxSDK. }
	\label{fig:2} 
\end{figure}

\paragraph*{End-to-end event-driven neuromorphic navigation system}
The physical implementation of the end-to-end event-driven neuromorphic navigation system is shown in Fig.~\ref{fig:2}. 
The system consists of an event camera (Prophesee EVK4), a neuromorphic processor (Intel Kapoho Bay), an edge computing platform (NVIDIA Jetson Orin NX), an x86-based host computer, and a micro aerial vehicle. 
Considering the limited on-chip neuron resources and the limited set of operators supported by Intel Kapoho Bay, we partition the overall spiking neural network into two modules for deployment: a perceptual representation module and a navigation module.
Specifically, the spiking neural networks for depth estimation and feature extraction (perceptual representation module)  are deployed on the NVIDIA Jetson Orin NX in the form of simulated SNNs, which generate event-based geometric representations of the environment. 
The spiking neural network for navigation (navigation module) is deployed on the Intel Kapoho Bay neuromorphic processor, enabling low-power and real-time action generation. 
Since Intel Kapoho Bay is not directly compatible with the ARM-based onboard platform, wireless communication is used to exchange data between the onboard platform and the host computer. 
Through the ROS communication framework, the perceptual representation module transmits states to the navigation module, while the navigation module returns the action for flight execution. 
In this way, a physical validation platform is established for evaluating the end-to-end event-driven neuromorphic navigation in real-world flight scenarios.

\begin{figure} 
	\centering
	\includegraphics[width=\textwidth]{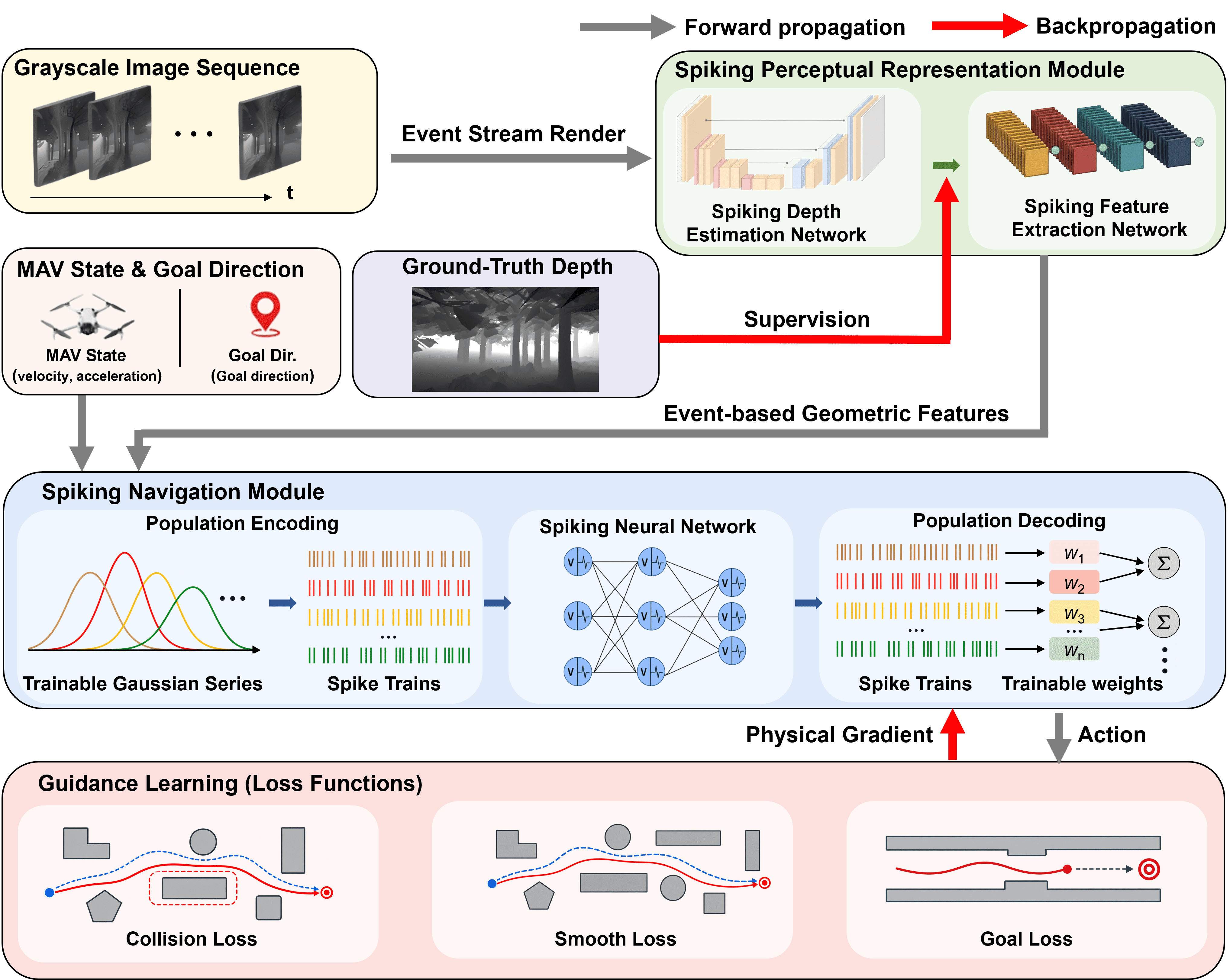}
	\caption{\textbf{Training pipeline of the proposed end-to-end event-driven neuromorphic navigation method.} Continuous grayscale image sequences are first transformed into asynchronous event streams using an event simulator. Depth estimation is then introduced as an auxiliary task to enable geometry-aware spiking features learning from event streams. The learned event features are fused with the MAV state and goal direction, and then encoded into spike trains via population coding. The resulting spikes are processed by the spiking navigation network and decoded to predict the action. Finally, collision loss, smoothness loss, and goal-reaching loss are defined based on the predicted action, providing physically meaningful supervision for end-to-end optimization of all network modules.}
	\label{fig:3} 
\end{figure}

To achieve end-to-end event-driven neuromorphic autonomous navigation, we construct the simulation-based training pipeline as illustrated in Fig.~\ref{fig:3}. 
First, grayscale image sequences are generated and converted into asynchronous event streams using an event simulator. 
The event streams are processed by a spiking depth estimation network and a spiking feature extraction network to learn geometry-aware environmental features. 
To bridge continuous navigation variables and spike-based computation, the extracted event features, robot state, and goal direction are encoded into spike trains through a multi-Gaussian population coding scheme with trainable mean and bias parameters. 
The outputs of the spiking navigation network are decoded into a continuous action through learnable population decoding weights. 
Given the predicted action, a polynomial trajectory parameterization module generates a sequence of candidate trajectory points, based on which differentiable losses are formulated to evaluate trajectory smoothness, collision risk, and target proximity. 
These losses provide physically meaningful supervision and allow gradients to be backpropagated through the entire pipeline, enabling joint optimization of the population encoding parameters, population decoding weights, spiking navigation network, and spiking perceptual representation network.
In parallel, the corresponding simulated depth ground truth is used as auxiliary supervision for the spiking depth estimation network. 
This auxiliary objective encourages geometrically consistent event-based representations, improving the robustness and generalization of the learned navigation policy.

\begin{figure} 
	\centering
	\includegraphics[width=\textwidth]{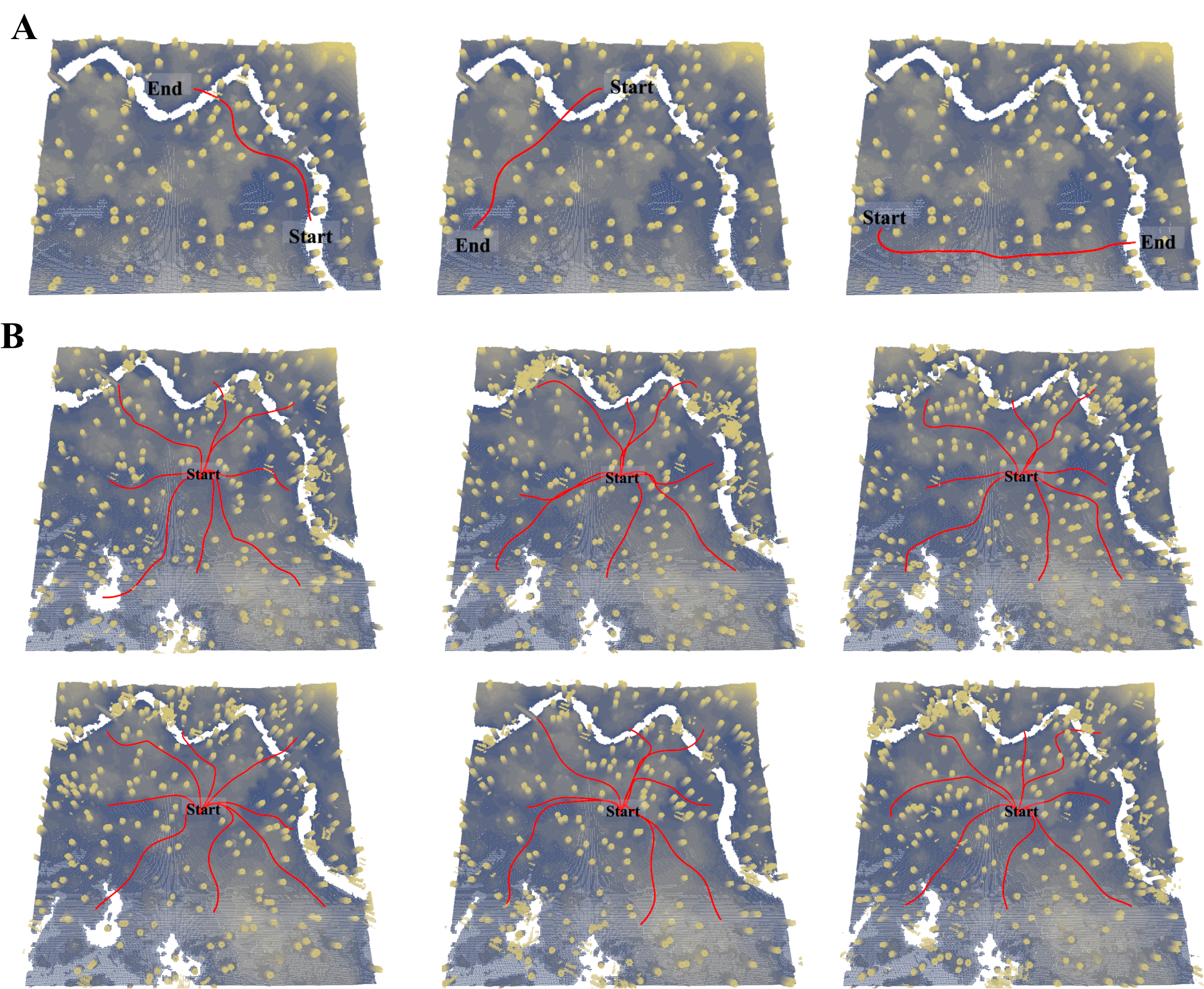}
	\caption{\textbf{Navigation trajectories of the proposed method in simulation environments. }  
    (A) Continuous navigation performance is evaluated by requiring the MAV to sequentially pass through multiple waypoints in a simulated environment. (B) Robustness is evaluated by placing target points in eight directions within each of six testing environments.}
	\label{fig:4} 
\end{figure}

\paragraph*{Navigation performance}
We first evaluate the navigation performance of the proposed method in the simulation environment. 
Specifically, the MAV is commanded to perform continuous autonomous flight in the simulated scene, and its trajectory is marked in red in the environment, as shown in Fig.~\ref{fig:4}A. 
The results show that the proposed end-to-end event-driven neuromorphic navigation method can generate continuous, smooth, and stable flight trajectories in the simulation environment, demonstrating its capability for continuous autonomous navigation.
To further evaluate the robustness of the proposed method across diverse environments, we design six testing environments, with eight target positions placed in each environment, as shown in Fig.~\ref{fig:4}B. 
The results indicate that the proposed method maintains robust navigation performance across different environments.

\begin{table}[htbp]
\begin{center}
\caption{\textbf{Quantitative comparison between the ANN-based method and the proposed method.} Navigation performance is evaluated at maximum flight speeds of 0.5, 1.0, and 2.0 m/s across six simulation environments. Success rate denotes the number of successful trials out of 48 total navigation trials, and average distance and average time denote the mean trajectory length and navigation time of the evaluated trials, respectively.}
\label{Table:1}
\vspace{2pt}
\setlength{\tabcolsep}{0.01\linewidth}
\begin{tabular}{ccccc}
\hline
\textbf{Scheme}
& \textbf{Max Speed (m/s)}
& \textbf{Success Rate}
& \textbf{Average Distance (m)}
& \textbf{Average Time (s)} \\
\hline

\begin{tabular}[c]{c}
ANN-based\\
Ours
\end{tabular}
&
0.5
&
\begin{tabular}[c]{c}
45/48\\
45/48
\end{tabular}
&
\begin{tabular}[c]{c}
23.35\\
24.35
\end{tabular}
&
\begin{tabular}[c]{c}
67.60\\
66.14
\end{tabular}
\\

\begin{tabular}[c]{c}
ANN-based\\
Ours
\end{tabular}
&
1.0
&
\begin{tabular}[c]{c}
47/48\\
46/48
\end{tabular}
&
\begin{tabular}[c]{c}
24.07\\
24.87
\end{tabular}
&
\begin{tabular}[c]{c}
33.87\\
32.45
\end{tabular}
\\

\begin{tabular}[c]{c}
ANN-based\\
Ours
\end{tabular}
&
2.0
&
\begin{tabular}[c]{c}
47/48\\
48/48
\end{tabular}
&
\begin{tabular}[c]{c}
24.82\\
25.98
\end{tabular}
&
\begin{tabular}[c]{c}
16.85\\
17.78
\end{tabular}
\\
\hline
\end{tabular}
\end{center}
\end{table}

\begin{figure} 
	\centering
	\includegraphics[width=0.85\textwidth]{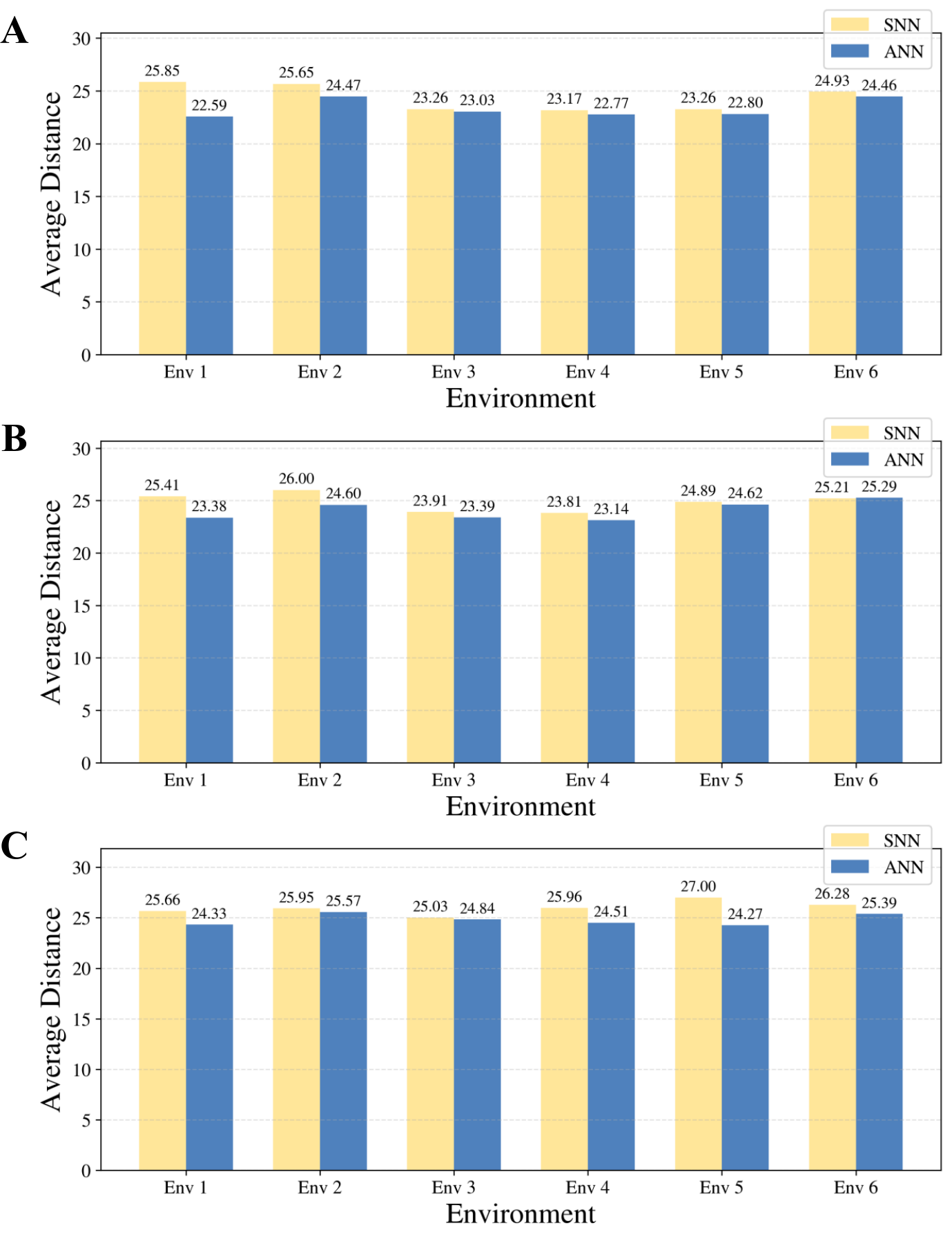}
	\caption{\textbf{Average trajectory distance comparison between our method and the ANN-based method across six environments.} (A) Average trajectory distance at a maximum speed of 0.5 m/s. (B) Average trajectory distance at a maximum speed of 1 m/s. (C) Average trajectory distance at a maximum speed of 2 m/s.}
	\label{fig:5} 
\end{figure}

\begin{figure} 
	\centering
	\includegraphics[width=0.85\textwidth]{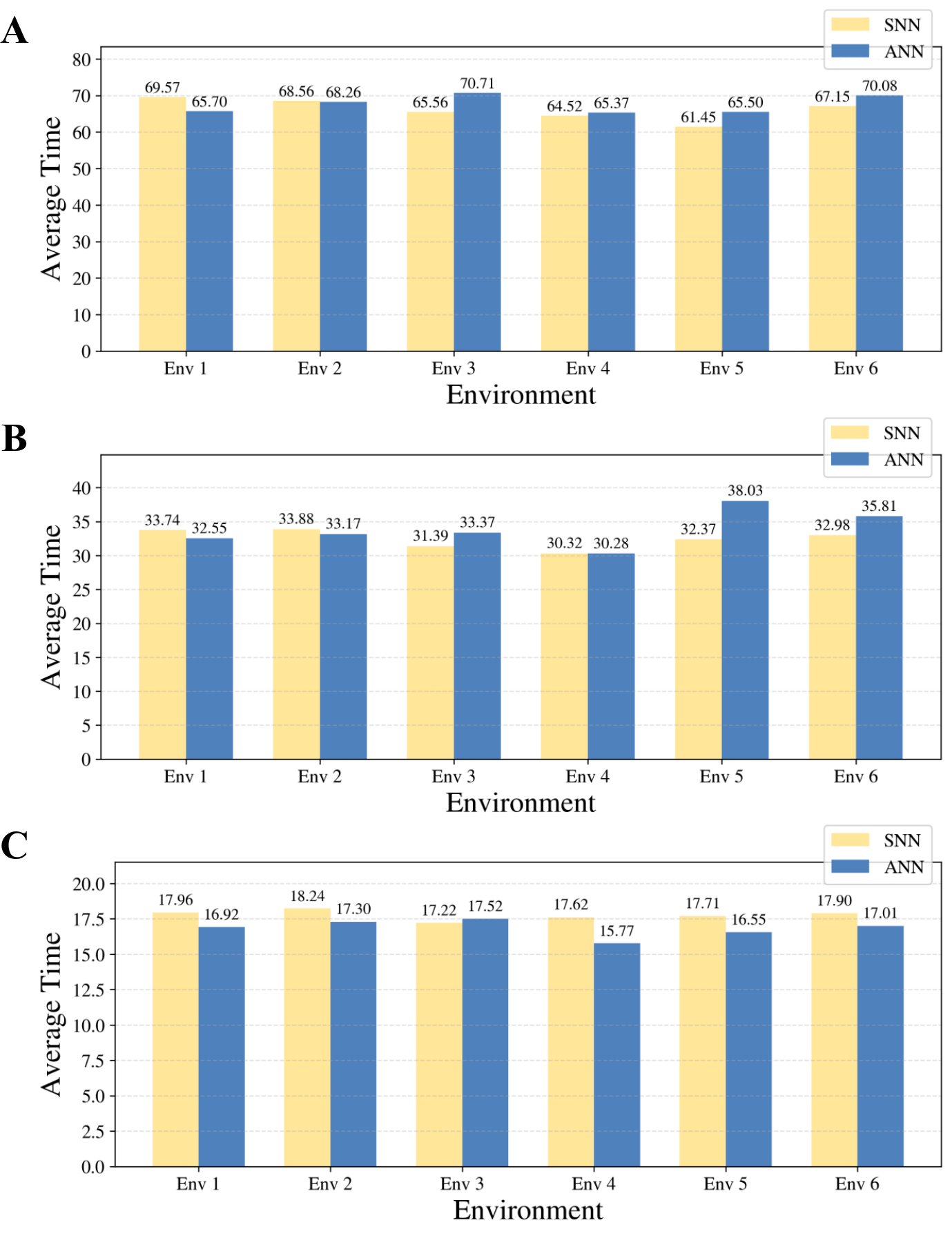}
	\caption{\textbf{Average navigation time comparison between our method and the ANN-based method across six environments.} (A) Average navigation time at a maximum speed of 0.5 m/s. (B) Average navigation time at a maximum speed of 1 m/s. (C) Average navigation time at a maximum speed of 2 m/s.}
	\label{fig:6} 
\end{figure}

\begin{figure} 
	\centering
	\includegraphics[width=\textwidth]{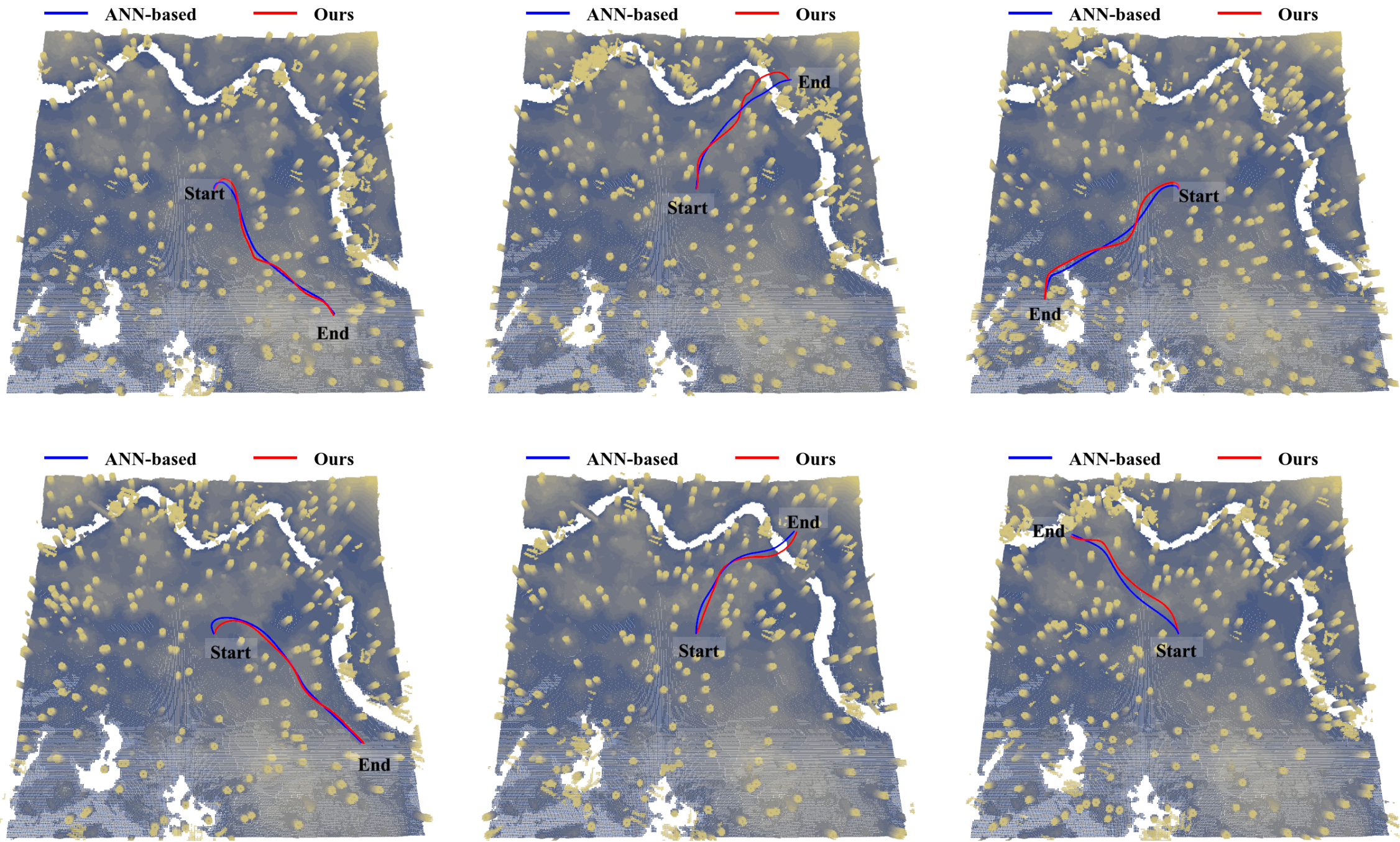}
	\caption{\textbf{Navigation trajectories comparison between the proposed method and the ANN-based method.} The navigation trajectories of the two methods are compared across six different simulation environments, with the trajectories of the proposed method shown in red and those of the ANN-based method shown in blue.}
	\label{fig:7} 
\end{figure}

\begin{figure} 
	\centering
	\includegraphics[width=\textwidth]{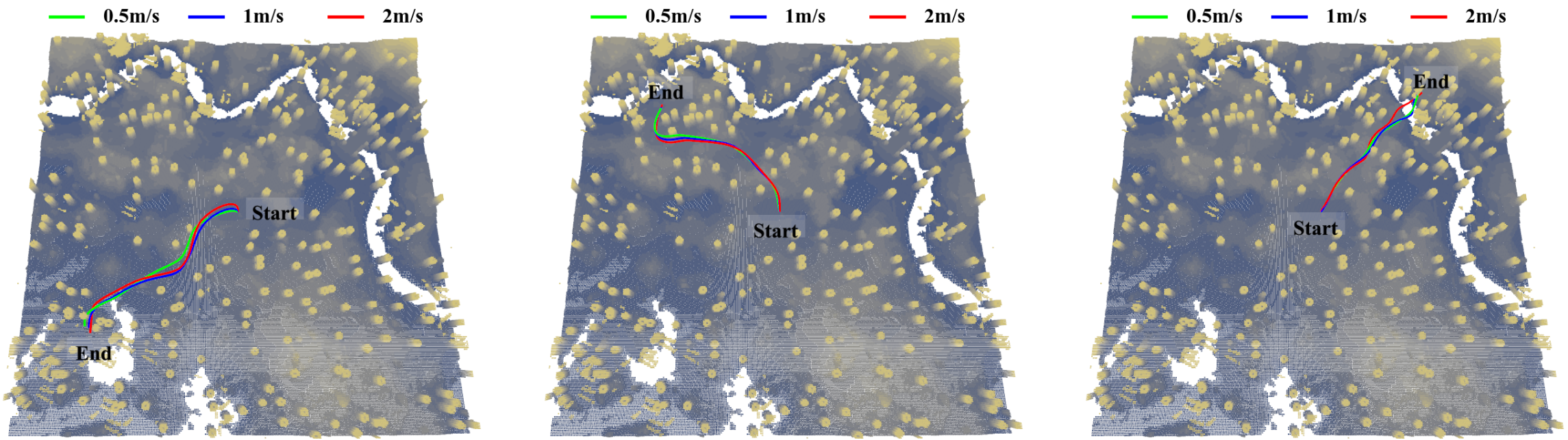}
	\caption{\textbf{Navigation trajectories at different maximum speeds.} The green, blue, and red lines indicate the navigation trajectories at maximum speeds of 0.5 m/s, 1.0 m/s, and 2.0 m/s, respectively.}
	\label{fig:8} 
\end{figure}

\begin{figure} 
	\centering
	\includegraphics[width=\textwidth]{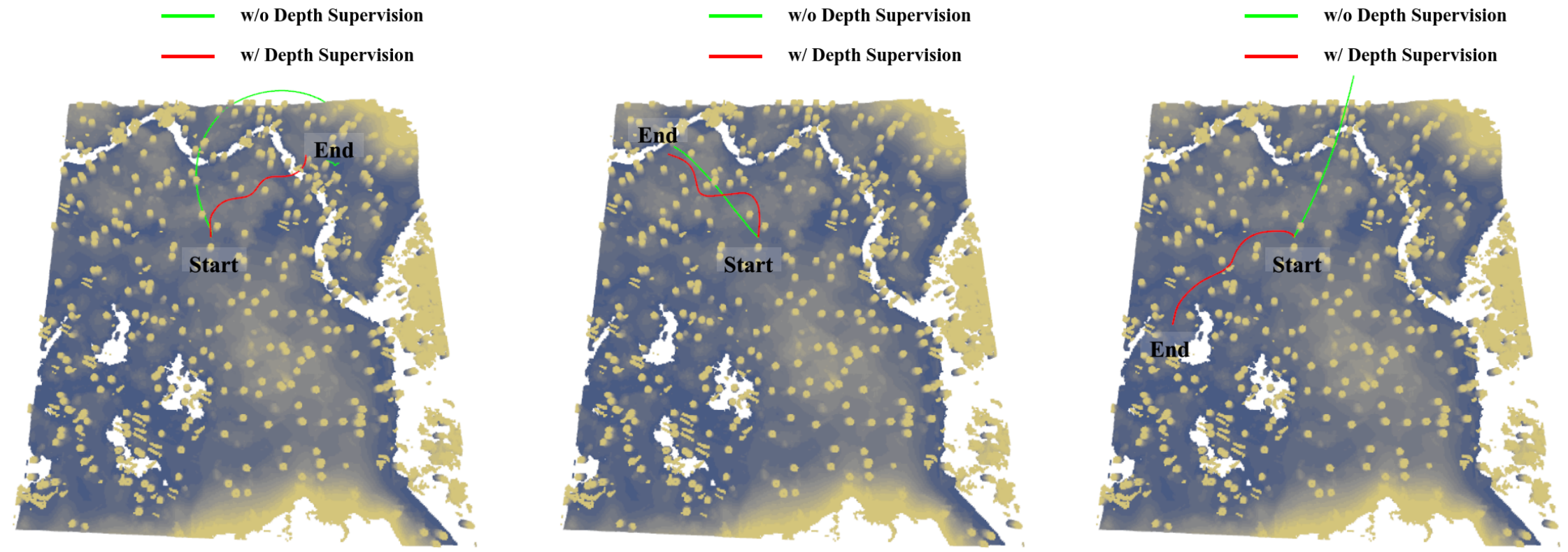}
	\caption{\textbf{Comparison of navigation trajectories with and without depth supervision.} The green and red lines indicate trajectories without and with depth supervision, respectively. Removing depth supervision leads to collisions, ground crashes, and substantial target deviations.}
	\label{fig:9} 
\end{figure}

We then compare the proposed SNN-based navigation method with its ANN-based counterpart.
Specifically, all SNN modules are replaced with their corresponding ANN modules, and both methods are evaluated under the same six environmental settings. 
To quantitatively evaluate navigation performance, we report the navigation success rate, average trajectory distance, and average navigation time across different maximum speeds.
Table~\ref{Table:1} summarizes the success rate, average trajectory distance, and navigation time across all tested environments, while Fig.~\ref{fig:5} and Fig.~\ref{fig:6} further break down the average trajectory distance and navigation time by environment at each maximum speed.
At the same maximum speed, the proposed method achieves performance comparable to the ANN-based method in terms of success rate, average trajectory distance, and average navigation time.
Fig.~\ref{fig:7} compares representative navigation trajectories of the proposed method with those of the ANN-based method. 
Taken together, these results demonstrate that, despite the challenges associated with training and optimizing SNNs for complex navigation tasks, the proposed neuromorphic method achieves navigation performance comparable to that of the ANN-based method while retaining spike-based computation.
Notably, the navigation success rate decreases at lower maximum speeds. 
This degradation is mainly attributed to the reduced event generation from distant objects during slow speed, which leads to less reliable depth estimation and consequently delays obstacle-avoidance responses, as illustrated in Fig.~\ref{fig:8}.
To further examine the role of depth supervision in navigation robustness, Fig.~\ref{fig:9} compares navigation trajectories with and without depth supervision. 
The red and green lines indicate trajectories with and without depth supervision, respectively.
Without depth supervision, the policy exhibits frequent collisions, ground contacts, and larger deviations from the target, whereas depth supervision yields more stable and goal-directed navigation.

\begin{figure} 
	\centering
	\includegraphics[width=\textwidth]{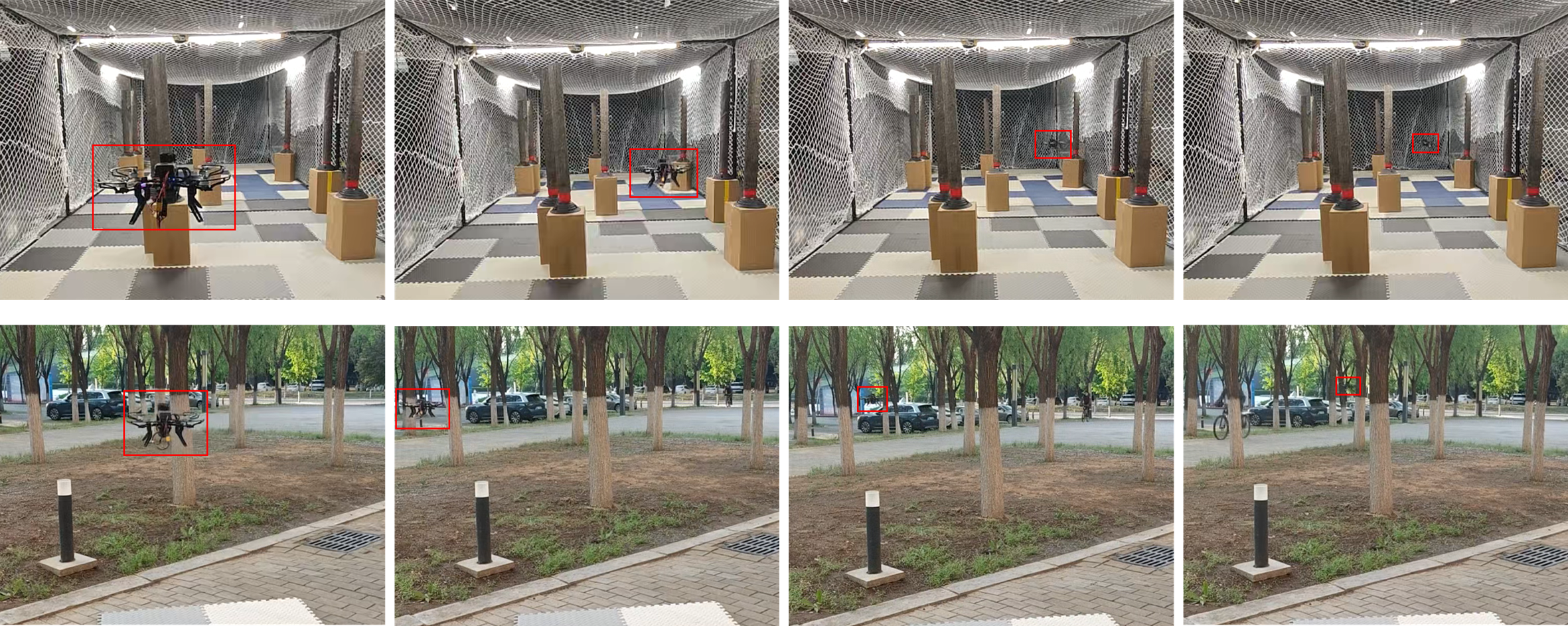}
	\caption{\textbf{Navigation performance of the proposed method in real-world environments.} Indoor and outdoor navigation tests were conducted by placing the target point behind a series of obstacles, requiring the MAV to avoid the obstacles and successfully reach the target.}
	\label{fig:10} 
\end{figure}

Finally, we deploy the proposed neuromorphic navigation method on a physical MAV and conduct both indoor and outdoor experiments, as shown in Fig.~\ref{fig:10}. 
In the indoor experiment, tree trunks are simulated using obstacles covered with artificial bark textures. 
The MAV takes off in front of multiple obstacles, with the target point placed behind them. 
We conducted ten real-world navigation trials in both indoor and outdoor environments. The proposed method successfully reached the target in seven out of ten indoor trials and nine out of ten outdoor trials. 
Compared with the simulation experiments, the navigation success rate decreased in both real-world settings, mainly due to two factors.
First, the lens equipped with the Prophesee EVK4 provides a relatively limited horizontal field of view, resulting in larger blind regions than those in simulation and restricting the MAV's perception of surrounding obstacles. 
Second, the depth maps captured by the Intel RealSense D435i are less reliable than the ideal depth information available in simulation, exhibiting more pronounced measurement noise and reduced accuracy for distant regions.
Notably, the quality of the indoor depth maps was substantially lower than that of the outdoor depth maps, which contributed to the more pronounced decrease in indoor navigation success rate. 
In some cases, although the event frames clearly captured the information of the environment, the limited quality of the depth supervision prevented the network from producing sufficiently accurate depth estimates, thereby affecting subsequent navigation decisions. 
In future work, we plan to employ a 3D LiDAR to provide more stable and accurate depth supervision for the event camera, with the aim of improving depth estimation accuracy in complex real-world environments.
Nevertheless, these real-world experiments demonstrate that the proposed neuromorphic navigation paradigm can be successfully transferred from simulation to physical flight.
Compared with previous neuromorphic MAV systems that focused primarily on single-axis control~\cite{vitale2021event} or optical-flow-based ego-motion control~\cite{paredes2024fully}, our system further extends the neuromorphic paradigm from low-level motion control to goal-directed autonomous navigation in complex environments.
To the best of our knowledge, this represents the first physical-system demonstration of end-to-end neuromorphic autonomous MAV navigation, highlighting the potential of the neuromorphic paradigm to support higher-level autonomous decision-making for real-world autonomous flight.

\paragraph*{Power and energy efficiency}
Compared with conventional artificial neural networks, spiking neural networks have inherent advantages in event-driven processing, sparse activation, and low-power computation, making them well suited for energy-efficient real-time autonomous navigation. 
To evaluate the computational efficiency of the proposed neuromorphic navigation method, we separately analyze the perceptual representation network implemented as a simulated SNN and the navigation network deployed on the Intel Kapoho Bay.

\begin{table}[htbp]
\begin{center}
\caption{\textbf{Per-inference energy consumption of SNN-based and ANN-based navigation networks across different devices.} Dynamic power is defined as the difference between the operating and idle power of each device. The inference rate denotes the number of forward inferences performed per second, and the energy per inference is calculated by dividing the dynamic power by the inference rate. The NVIDIA Jetson Orin NX is evaluated in both high-performance mode (25 W) and power-saving mode (10 W).}
\vspace{4pt}

\setlength{\tabcolsep}{0.005\linewidth}

\begin{tabular}{ccccc}
\hline
\textbf{Device}
& \textbf{Scheme}
& \begin{tabular}{c}
\textbf{Dynamic Power}\\
\textbf{(mW)}
\end{tabular}
& \begin{tabular}{c}
\textbf{Inference Rate}\\
\textbf{(s$^{-1}$)}
\end{tabular}
& \begin{tabular}{c}
\textbf{Energy per Inference}\\
\textbf{(mJ/inf.)}
\end{tabular} \\
\hline

\begin{tabular}{c}
NVIDIA GeForce\\
RTX 4090
\end{tabular}
& ANN-based
& 43000
& 2078
& 20.69 \\

\begin{tabular}{c}
NVIDIA Jetson Orin \\
NX (25 W)
\end{tabular}
& ANN-based
& 5275
& 805
& 6.55 \\

\begin{tabular}{c}
NVIDIA Jetson Orin\\ 
NX (10 W)
\end{tabular}
& ANN-based
& 1958
& 649
& 3.02 \\

\begin{tabular}{c}
Intel Kapoho\\
Bay
\end{tabular}
& Ours
& 140
& 172
& 0.81 \\

\hline
\end{tabular}
\label{Table:2}
\end{center}
\end{table}

\begin{figure} 
	\centering
	\includegraphics[width=\textwidth]{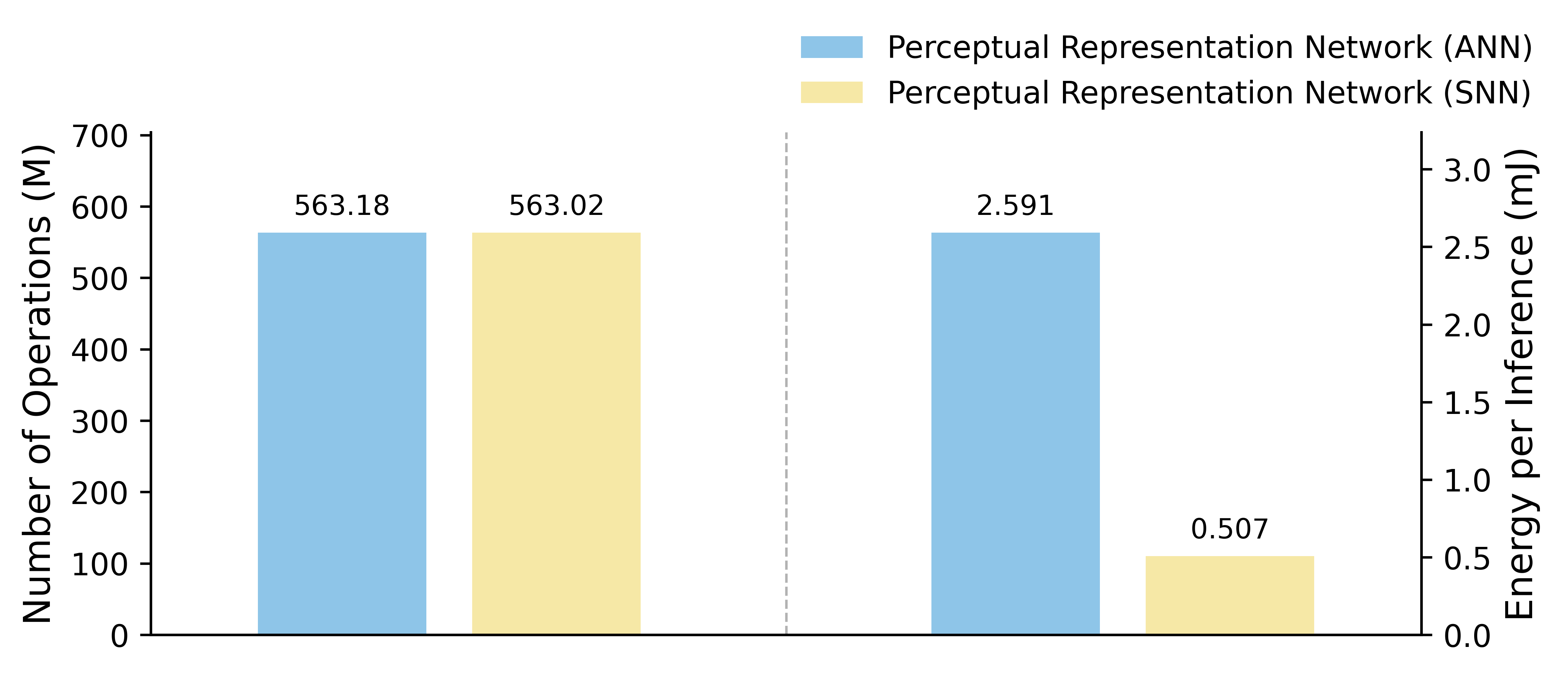}
	\caption{\textbf{Operations and per-inference energy consumption of ANN-based and SNN-based perceptual representation networks.} The ANN-based network is evaluated based on multiply–accumulate operations (MACs), whereas the SNN-based network is evaluated based on spike-triggered accumulate operations (ACs). Per-inference energy consumption is estimated from the corresponding operation counts using energy costs of 4.6 pJ per MAC and 0.9 pJ per AC under a 45-nm CMOS technology assumption.}
	\label{fig:11} 
\end{figure}

To estimate the energy consumption of the perceptual representation network, we follow the methodology of previous studies~\cite{su2023deep, hu2024advancing} by counting the number of accumulate operations (ACs) in the SNN and the number of multiply accumulate operations (MACs) in its ANN-based counterpart. 
The energy consumption is then calculated based on the operation counts under a 45-nm CMOS technology assumption~\cite{horowitz20141}, where each MAC and AC operation consumes 4.6 pJ and 0.9 pJ, respectively. 
The results are shown in Fig.~\ref{fig:11}. 
Under the same network architecture, the ANN-based perceptual representation network requires 563.18M MAC operations and consumes 2.591 mJ per inference. 
In contrast, the SNN-based network requires 563.02M AC operations and consumes only 0.507 mJ per inference. 
These results indicate that the SNN-based perceptual representation network achieves substantially lower inference energy, requiring only 19.57\% of the per-inference energy consumed by its ANN-based counterpart.

For the navigation network deployed on the neuromorphic processor, we further compare its energy consumption with that of the corresponding ANN-based model deployed on conventional ANN computing platforms. 
Three hardware platforms are considered: NVIDIA GeForce RTX 4090, NVIDIA Jetson Orin NX, and Intel Kapoho Bay neuromorphic processor. 
The ANN-based models are deployed on the NVIDIA GeForce RTX 4090 and NVIDIA Jetson Orin NX, whereas the SNN-based model is deployed on Intel Kapoho Bay. 
To quantify the per-inference energy consumption of different deployment schemes, we measure the dynamic power consumption and inference speed of each model on its corresponding hardware platform, as reported in Table~\ref{Table:2}.
Here, dynamic power is defined as the difference between the operating power and the idle power of the device. 
For the NVIDIA Jetson Orin NX, both the high-performance mode (25 W) and the power-saving mode (10 W) are evaluated.
The experimental results show that the per-inference energy consumption of the ANN-based model deployed on the NVIDIA GeForce RTX 4090 is approximately 25.54 times that of the SNN-based model deployed on Intel Kapoho Bay. 
For the NVIDIA Jetson Orin NX edge computing platform, the per-inference energy consumption of the ANN-based model is approximately 8.09 times that of the SNN-based model in the high-performance mode (25 W).
Even in the power-saving mode (10 W), its per-inference energy consumption remains approximately 3.73 times that of the SNN-based model. 
These results demonstrate the substantial inference energy-efficiency advantage of neuromorphic deployment, with the spiking navigation network running on Intel Kapoho Bay consuming markedly less energy than its ANN-based counterpart deployed on conventional computing platforms.

\subsection*{Discussion}
In this paper, we have presented a neuromorphic visual navigation paradigm for micro aerial vehicles, inspired by the asynchronous, sparse, and event-driven sensing and computation principles of biological visual systems. 
Unlike conventional MAV navigation systems that rely predominantly on frame-based cameras and artificial neural networks, the proposed paradigm integrates event-based sensing, spiking neural computation, and neuromorphic hardware within a unified autonomous navigation framework.
By leveraging sparse and asynchronous information acquisition and processing inspired by biological visual systems, the proposed paradigm provides a promising pathway toward more efficient and robust MAV visual navigation.

The proposed method addresses several key bottlenecks in translating neuromorphic navigation from algorithmic design to practical robotic operation. 
A spiking perceptual representation network extracts obstacle-relevant geometric features from the event stream, compensating for the lack of explicit spatial structure in raw events. 
Population coding and guidance learning further enable the spiking navigation network to efficiently learn navigation policies within a limited number of time steps. 
Beyond algorithmic development, we establish a complete pipeline from simulation-based training to physical deployment, integrating neuromorphic perception, spiking decision-making, and hardware execution within a physical MAV system. 
Experimental results demonstrate navigation performance comparable to that of the ANN counterpart while substantially reducing inference energy consumption, and real-world flight experiments further validate the feasibility of the complete neuromorphic navigation framework.

Despite these results, the current physical system still has two main limitations. 
First, owing to the limited neuron models and capacity supported by the Intel Kapoho Bay platform, full onboard deployment of all spiking neural networks on neuromorphic hardware has not yet been achieved, and the system therefore still relies on additional wireless data transmission. 
Second, because existing monocular event-based visual odometry remains limited in estimation accuracy and computational cost, pose estimation still depends on a conventional frame-based camera. 
Although the event camera provides more robust visual perception during rapid motion and under challenging illumination, the overall robustness of the system remains constrained by the external frame-based pose estimation module. 
Future work will explore more capable neuromorphic hardware with greater capacity to enable more complete onboard neuromorphic deployment, while also developing lightweight and accurate event-based visual odometry to extend event-based perception to state estimation and ultimately achieve a more complete neuromorphic navigation loop.

Collectively, \textit{this work moves neuromorphic MAV navigation beyond isolated sensing or computation modules toward a unified perception–decision–execution loop.} 
The successful physical-system validation demonstrates that the asynchronous, sparse, and event-driven principles underlying biological information processing can be translated into an operational robotic navigation system rather than remaining confined to individual neuromorphic components or simulation. 
By bringing neuromorphic sensing, spiking computation, and hardware execution together within an embodied autonomous platform, this framework provides a viable path toward robust and low-power robotic systems that more closely exploit the computational principles of biological intelligence.

\subsection*{Materials and Methods}
This section presents the complete pipeline of the end-to-end event-driven neuromorphic navigation method. 
We first introduce event cameras and spiking neural networks, which form the sensing and computing foundations of our framework. 
We then describe the main components of the proposed method, including event-based geometric representation, population coding for spike trains encoding and decoding, state and action spaces, and the end-to-end training procedure of the navigation policy. 
Next, we present the detailed procedure for power consumption measurement, including operation-count-based energy estimation and hardware-based dynamic power measurement. Finally, we describe the hardware deployment of the proposed method.

\paragraph*{Event cameras} Event cameras are neuromorphic vision sensors inspired by the biological retina. 
Unlike conventional frame-based cameras, they asynchronously respond to changes in luminance at each pixel and generate sparse event streams in the address-event representation (AER) format.
Each pixel operates independently and triggers an event when the change in logarithmic luminance exceeds a predefined contrast threshold.
The change in brightness at a pixel is defined as:
\begin{equation}
\label{eq:1}
\begin{aligned}
\Delta \boldsymbol{L}\left(x_{k}, y_{k}, t_{k}\right)&=\boldsymbol{L}\left(x_{k}, y_{k}, t_{k}\right)-\boldsymbol{L}\left(x_{k}, y_{k}, t_{k}-\Delta t_{k}\right),
\end{aligned}
\end{equation}
where $L(\cdot)$ denotes the logarithmic luminance and $\Delta t_{k}$ is the time interval since the previous event at the same pixel.
When the luminance change exceeds the contrast threshold, the corresponding pixel generates an event. 
The event polarity $p_{k}$ is determined by the sign of the luminance change:
\begin{equation}
\label{eq:2}
    p_{k}=
    \begin{cases}
    1   &\text{if } \Delta \boldsymbol{L}\left(x_{k}, y_{k}, t_{k}\right)>\vartheta\\
    -1   &\text{if } \Delta \boldsymbol{L}\left(x_{k}, y_{k}, t_{k}\right)<-\vartheta
    \end{cases},
\end{equation}
where $\vartheta$ denotes the contrast threshold. 
Each event is therefore represented by its pixel coordinates, timestamp, and polarity as
$\boldsymbol{e}_{k}=\left(x_{k}, y_{k}, t_{k}, p_{k}\right)^{\top}$.

\paragraph*{Spiking neural networks} 
We construct the spiking neural networks using leaky integrate-and-fire (LIF) neurons. 
Two types of LIF neurons are considered in this work: the current-based LIF (CUBA-LIF) neuron and the dual-state LIF (TS-LIF) neuron. 
The CUBA-LIF neuron models neuronal dynamics using a single membrane-potential state, whereas the TS-LIF neuron explicitly separates the synaptic current state from the membrane-potential state, allowing independent modeling of current and voltage decay.
For the CUBA-LIF neuron model, the iterative update equations for the $n$-th layer of the SNN are defined as:
\begin{equation}
\label{eq:3}
\begin{aligned} 
\boldsymbol{u}_{n}^{t} &=(1-\frac{1}{\tau}) \boldsymbol{u}_{n}^{t-1} \boldsymbol{g}\left(\boldsymbol{s}_{n}^{t-1}\right)+\frac{1}{\tau}\boldsymbol{W}_{n}\boldsymbol{s}_{n}^{t}    \\ 
\boldsymbol{s}_{n+1}^{t} &=H(\boldsymbol{u}_{n}^{t}-u_{\mathrm{th}})
\end{aligned},
\end{equation}
For the TS-LIF neuron model, the iterative update equations can be described as follows:
\begin{equation}
\label{eq:4}
\begin{aligned} 
\boldsymbol{c}_{n}^{t} &=\delta_{\text{curr}} \boldsymbol{c}_{n}^{t-1} + \boldsymbol{W}_{n} \boldsymbol{s}_{n}^{t} \\
\boldsymbol{u}_{n}^{t} &=\delta_{\text{volt}} \boldsymbol{u}_{n}^{t-1} \boldsymbol{g}\left(\boldsymbol{s}_{n}^{t-1}\right)+\boldsymbol{c}_{n}^{t}    \\ 
\boldsymbol{s}_{n+1}^{t} &=H(\boldsymbol{u}_{n}^{t}-u_{\mathrm{th}})
\end{aligned},
\end{equation}
Here, $t \in[1, \mathrm{T}]$ denotes discrete time step, 
$\boldsymbol{W}_n$ is the input weight matrix, $\boldsymbol{c}_{n}^{t}$ and $\boldsymbol{u}_{n}^{t}$ represent the synaptic current and membrane potential. 
$\boldsymbol{s}_{n}^{t}$ denotes the input spike trains of the $n$-th layer, while $\boldsymbol{s}_{n+1}^{t}$ denotes the output spike trains.
$\tau$ is the membrane time constant. 
$\delta_{\text{curr}} $ and $\delta_{\text{volt}}$ are the decay coefficients of the synaptic current and membrane potential, respectively.
$\boldsymbol{g}\left(\boldsymbol{s}_{n}^{t-1}\right)=\mathbf{1}-\boldsymbol{s}_{n}^{t-1}$ is the reset function. 
$H$ represents the Heaviside function.
$u_{\mathrm{th}}$ is the firing threshold.

\paragraph*{Event-based geometric representation}
Raw event streams mainly encode asynchronous brightness changes and lack explicit geometric structure, making it difficult for the navigation policy to infer obstacle layouts and learn effective avoidance behaviors. 
To address this issue, we introduce event-based geometric representation learning, where a spiking depth estimation network first predicts scene depth from event frames, and a spiking feature extraction network further extracts compact geometry-aware features for navigation.
Given an event stream within a temporal window $[t_{\mathrm{b}}, t_{\mathrm{f}}]$, we first accumulate events at each pixel location to construct an event frame:
\begin{equation}
\label{eq:5}
E(x_i,y_i)=\sum_{t=t_{\text{b}}}^{t_{\text{f}}}\boldsymbol{e}(x_i,y_i),
\end{equation}
where $t_{\text{b}}$ and $t_{\text{f}}$ represent the start and end times of the interval, respectively.
Considering event polarity, the event frame is further represented as a two-channel tensor $E_{\mathrm{tc}}$, where the two channels correspond to positive and negative events.
We employ a spiking neural network-based U-Net to perform depth estimation, where the membrane potentials of the final layer are used as the estimated depth values.
The spiking U-Net consists of four encoder layers, four decoder layers, one bottleneck layer, and a depth prediction head. 
Each encoder and decoder module contains two convolutional units for hierarchical event feature extraction and reconstruction. 
In the decoding stage, each decoder module is preceded by an upsampling module, which restores the feature resolution and aligns it with the corresponding skip-connected feature from the encoder.
The predicted depth map $D$ is expressed as follows:
\begin{equation}
\label{eq:6}
    D=f_{\text{SNN-UNET}}(E_{\text{tc}})
\end{equation}
Subsequently, a spiking residual network is adopted to further encode the estimated depth map into compact geometric feature representations. 
The network is built upon the SEW-ResNet architecture~\cite{fang2021deep} and consists of a stem module, two residual layers, and a feature output head. 
Each residual layer contains two residual blocks, enabling hierarchical extraction of navigation-relevant geometric features from the estimated depth representation. 
The final event-based geometric features are given by:
\begin{equation}
\label{eq:7}
    \boldsymbol{Z}=f_{\text{SEW}}(D)
\end{equation}

\paragraph*{Population coding for spike trains} 
To enable accurate representation of continuous-valued states within a limited number of time steps, we employ a population-coding mechanism for both encoding and decoding in the spiking neural network (SNN).
For each input state $\xi_i$, $N_{\mathrm{in}}$ Gaussian receptive fields are used for encoding.
Both the mean $\mu_n$ and the standard deviation $\sigma_n$ of each Gaussian receptive field are learnable parameters.
The standard deviations $\sigma_n$ are initialized to a fixed value while the means are initialized according to the distribution range of the input state $\boldsymbol{\xi}$:
\begin{equation}
\label{eq:8}
    \mu_{n}=d_{\mathrm{min}}+\frac{n-1}{N_{\mathrm{in}}-1}(d_{\mathrm{max}}-d_{\mathrm{min}}),
\end{equation}
where $d_{\mathrm{min}}$ and $d_{\mathrm{max}}$ denote the minimum and maximum values of the input $\boldsymbol{\xi}$, respectively.
Given the Gaussian receptive fields, each input $\xi_i$ is transformed into a set of activation intensities $I_{i,n}$:
\begin{equation}
\label{eq:9}
    I_{i,n}=e^{-\frac{1}{2}(\frac{\xi_{i}-\mu_{n}}{\sigma})^{2}}
\end{equation}
where $\mu_n$ and $\sigma_n$ denote the learnable mean and standard deviation of the $n$-th Gaussian receptive field, respectively.
A deterministic spike-generation process is then applied to convert the activation intensities into input spike trains $\boldsymbol{O}_{\textrm{in},t}$: 
\begin{equation}
\label{eq:10}
\begin{aligned} 
\boldsymbol{v}_{t}&=\boldsymbol{v}_{t-1} + \boldsymbol{I}-\boldsymbol{O}_{\textrm{in},t-1}u_{\text{th,en}} \\
\boldsymbol{O}_{\textrm{in},t} &=\boldsymbol{H}(\boldsymbol{v}_{t}-u_{\text{th, en}})
\end{aligned},
\end{equation}
where $u_{\text{th, en}}$ denotes the firing threshold for spike encoding.
After population encoding, each state $s_i$ is represented jointly by $N_{\mathrm{in}}$ spike trains.
To accurately regress the action, we further employ a population decoding strategy.
In this strategy, $N_{\mathrm{out}}$ output neurons jointly represent a single continuous output value.
Specifically, the firing rate $r_i$ of the $i$-th output spike train $\boldsymbol{O}_{i}^{t}$
over the time window $\mathrm{T}$ is first calculated as:
\begin{equation}
\label{eq:11}
    r_{i}=\frac{1}{\mathrm{T}} \sum_{t=1}^{\mathrm{T}} \boldsymbol{O}_{i}^{t}
\end{equation}
The action prediction is then obtained through a trainable linear readout over the output firing rates:
\begin{equation}
\label{eq:12}  a=\sum_{i=1}^{N_{\mathrm{op}}}\boldsymbol{w}_{p,i}r_{i}+\boldsymbol{b}_{p,i}
\end{equation}
where $\boldsymbol{w}_{p}$ and $\boldsymbol{b}_{p}$ denote the trainable readout weights and biases, respectively.

\paragraph*{State and action spaces}
Here, we first define the state and action spaces of the spiking navigation network.
At each inference step, the spiking navigation network receives an input state $\boldsymbol{\zeta}_t$, which consists of the normalized velocity $\bar{\boldsymbol{v}}_t$, acceleration $\bar{\boldsymbol{a}}_t$, goal direction $\bar{\boldsymbol{g}}_t$ in the camera frame, and event-based visual features $\boldsymbol{Z}$.
\begin{equation}
\label{eq:13} 
\boldsymbol{\zeta}_t
=
[
\bar{\boldsymbol{v}}_t,
\bar{\boldsymbol{a}}_t,
\bar{\boldsymbol{g}}_t,
\boldsymbol{Z}
].
\end{equation}
where goal direction is defined as follows:
\begin{equation}
\label{eq:14} 
    \bar{\boldsymbol{g}}_t
    =
    \frac{
    \boldsymbol{p}_{\mathrm{goal}}
    -
    \boldsymbol{p}_{\mathrm{t}}
    }{
    \left\|
    \boldsymbol{p}_{\mathrm{goal}}
    -
    \boldsymbol{p}_{\mathrm{t}}
    \right\|_2
    },
\end{equation}
where $\boldsymbol{p}_{\mathrm{goal}}$ and $\boldsymbol{p}_{\mathrm{t}}$ denote the pose of the goal and the MAV, $|\cdot|_2$ denotes the Euclidean norm.
We then generate a primitive waypoint $p$ in the camera frame using polar coordinates:
\begin{equation}
\label{eq:15} 
    p=(r\cos\alpha\cos\theta, r\sin\alpha\cos\theta, r\sin\theta)
\end{equation}
where $r$, $\alpha$, and $\theta$ denote the radial distance, azimuth angle, and polar angle, respectively.
The spiking navigation network predicts the offsets $[\Delta r, \Delta \alpha, \Delta \theta]$, velocities $\boldsymbol{v}=[v_{x}, v_{y}, v_{z}]$ and accelerations $\boldsymbol{a}=[a_{x}, a_{y}, a_{z}]$ of the waypoint.
Then, the waypoint $\tilde{p}$ can be expressed as:
\begin{equation}
\label{eq:16} \tilde{p}=(\tilde{r}\cos\tilde{\alpha}\cos\tilde{\theta}, \tilde{r}\sin\tilde{\alpha}\cos\tilde{\theta}, \tilde{r}\sin\tilde{\theta})
\end{equation}
where $\tilde{r}=r+\Delta r$, $\tilde{\alpha}=\alpha+\Delta \alpha$ and $\tilde{\theta}=\theta+\Delta \theta$.
Then, the end state $\boldsymbol{\zeta}_{\text{end}}$ can be expressed as:
\begin{equation}
\label{eq:17} 
    \boldsymbol{\zeta}_{\text{end}}=[\tilde{p}, \boldsymbol{v}, \boldsymbol{a}]
\end{equation}

\paragraph*{Training procedure for the neuromorphic visual navigation policy}
The network is trained with two supervision objectives, namely the geometric loss $\mathcal{L}_{\mathrm{geo}}$ and the trajectory loss $\mathcal{L}_{\mathrm{tra}}$. 
The geometric loss is used to enforce consistency between the network-predicted depth and the ground-truth depth, thereby improving the accuracy of geometric scene representation:
\begin{equation}
\label{eq:18}
\mathcal{L}_{\mathrm{geo}} = \ell_{\mathrm{s1}}(\hat{D}, D)
\end{equation}
where $\ell_{\mathrm{s1}}$ denotes the smooth L1 loss$, \hat{D}$ is the predicted depth, and $D$ is the ground-truth depth.
At the trajectory-supervision level, guidance learning~\cite{Yang-RSS-23,lu2024you} is employed to provide physically meaningful gradients for navigation policy optimization.
The trajectory loss $\mathcal{L}_{\mathrm{tra}}$ consists of three terms: the smoothness loss $\mathcal{L}_{s}$, the safety loss $\mathcal{L}_{o}$, and the goal loss $\mathcal{L}_{g}$:
\begin{equation}
\label{eq:19}
\mathcal{L}_{\mathrm{tra}}=w_{s}\mathcal{L}_{s}+w_{o}\mathcal{L}_{o}+w_{g}\mathcal{L}_{g}
\end{equation}
where $w_{s}$, $w_{o}$ and $w_{g}$ denote the corresponding weighting coefficients.
These terms guide the policy toward trajectories that are smooth, safe, and consistent with the navigation goal.
To compute the trajectory losses and their gradients, we parameterize the trajectory between the initial state $\boldsymbol{\zeta}_{\mathrm{initial}}$ and the end state $\boldsymbol{\zeta}_{\mathrm{end}}$ using a quintic polynomial.
For each spatial dimension, the trajectory is expressed as:
\begin{equation}
\label{eq:20}
p(t)=a_{0}+a_{1}t+a_{2}t^{2}+a_{3}t^{3}+a_{4}t^{4}+a_{5}t^{5}
\end{equation}
Let $A$ denote the coefficient vector of the polynomial.
The boundary condition vector $b$ is determined by the initial and end states:
\begin{equation}
\label{eq:21}
    b=[\zeta_{\mathrm{initial}},\zeta_{\mathrm{end}}]^{\top}
\end{equation}
A mapping matrix $M$ is then constructed to map the boundary conditions $b$ to the polynomial coefficients $A$:
\begin{equation}
\label{eq:22}
    A=Mb
\end{equation}
The smoothness loss is defined as:
\begin{equation}
\label{eq:23}
    \mathcal{L}_{s}=A^{\top}HA
\end{equation}
where $H$ denotes the smoothness cost matrix in polynomial coefficient space.
Accordingly, the Eq.~\eqref{eq:23} can be rewritten as:
\begin{equation}
\label{eq:24}
    \mathcal{L}_{s}=[\zeta_{\mathrm{initial}},\zeta_{\mathrm{end}}]M^{\top}HM[\zeta_{\mathrm{initial}},\zeta_{\mathrm{end}}]^{\top}
\end{equation}
Denote $M^{\top}HM$ as matrix B, the Eq.~\eqref{eq:24} can be further rewritten as:
\begin{equation}
\label{eq:25}
\mathcal{L}_{s}
=
\begin{bmatrix}
\zeta_{\mathrm{initial}} & \zeta_{\mathrm{end}}
\end{bmatrix}
\begin{bmatrix}
B_{\mathrm{ii}} & B_{\mathrm{ie}} \\
B_{\mathrm{ei}} & B_{\mathrm{ee}}
\end{bmatrix}
\begin{bmatrix}
\zeta_{\mathrm{initial}} \\
\zeta_{\mathrm{end}}
\end{bmatrix}
\end{equation}
Based on this quadratic form, the gradient of the smoothness loss can be computed as follows:
\begin{equation}
\label{eq:26}
\nabla_{\zeta_{\mathrm{end}}}\mathcal{L}_{s}=2\zeta_{\mathrm{initial}}^{\top}B_{\mathrm{ie}}+2\zeta_{\mathrm{end}}^{\top}B_{\mathrm{ee}}
\end{equation}
For obstacle avoidance, we define a distance penalty function as:
\begin{equation}
\label{eq:27}
    f_{\mathrm{dp}}(p(t))=e^{-\frac{d(t)-d_{0}}{\kappa}}
\end{equation}
where $d(t)$ denotes the distance from $p(t)$
to the nearest obstacle, $d_{0}$ is the safety threshold and $\kappa$ is the scaling factor.
Then, we discretize the total trajectory time $\mathrm{T}$ into k steps:
\begin{equation}
\label{eq:28}
    \mathrm{T}=K\Delta t
\end{equation}
Here, $\mathrm{T}$ is defined as $\mathrm{T}=\frac{\mathrm{C}}{v_{\text{max}}}$, where $\mathrm{C}$ is a constant and $v_{\text{max}}$ denotes the maximum speed.
The safety loss $\mathcal{L}_{o}$ can be expressed as:
\begin{equation}
\label{eq:29}
   \mathcal{L}_{o}=\sum_{k=0}^Kf_{\mathrm{dp}}(p(k\Delta t))\Delta t
\end{equation}
The trajectory point at the discretized time $k\Delta t$ can be written as:
\begin{equation}
\label{eq:30}
    p(k\Delta t)= \Gamma(k\Delta t)Mb
\end{equation}
where $\Gamma(k\Delta t)=[1,k\Delta t,(k\Delta t)^2,(k\Delta t)^3,(k\Delta t)^4,(k\Delta t)^5]$. 
The safety gradient can be calculated as:
\begin{equation}
\label{eq:31}
\nabla_{\zeta_{\mathrm{end}}}\mathcal{L}_{o}=\sum_{m=0}^k \left.\frac{\partial f_{\mathrm{dp}}(p(t))}{\partial p(t)}\frac{\partial p(t)}{\partial \zeta_{\mathrm{end}}}\right|_{t=k\Delta t}\Delta t
\end{equation}
Based on the above formulation, the gradient of the safety loss can be computed as follows:
\begin{equation}
\label{eq:32}
\nabla_{\zeta_{\mathrm{end}}}\mathcal{L}_{o}=\sum_{k=0}^K \nabla_{p}f_{\mathrm{dp}}(k\Delta t)\Gamma(k\Delta t)M_{\mathrm{end}}\Delta t
\end{equation}
where $M_{\mathrm{end}}$ represents the partial block of the matrix $M$ which corresponds to the end state $\zeta_{\mathrm{end}}$.
We define the goal loss as:
\begin{equation}
\label{eq:33}
    \mathcal{L}_{g}=(\mathbf{\zeta}_{\mathrm{end}, p}-p_{\mathrm{goal}})^{2}
\end{equation}
where $\mathbf{\zeta}_{\mathrm{end},p}$ represents the position component of the end state $s_{\mathrm{end}}$ and $p_{\mathrm{goal}}$ represents the goal position.
Then the gradient of the goal loss can be computed as follows:
\begin{equation}
\label{eq:34}
\nabla_{\zeta_{\mathrm{end}}}\mathcal{L}_{g}=2(\mathbf{\zeta}_{\mathrm{end},p}-p_{\mathrm{goal}})
\end{equation}
The geometric loss, trajectory loss, and corresponding guidance gradients are used to optimize the spiking navigation policy.
Since the spike-generation function is non-differentiable, the surrogate gradient method~\cite{wu2018spatio, fang2023spikingjelly} is adopted to enable backpropagation through the spiking neural network.

\paragraph*{Power and energy measurements}
For networks running in simulation, we estimate the per-inference energy consumption by counting the number of computational operations and combining it with the energy cost of each operation. 
For artificial neural networks (ANNs), forward inference is composed of floating-point multiply–accumulate operations.
Therefore, we count the number of multiply–accumulate operations (MACs) in each layer as the computational operations of the ANN. 
In contrast, spiking neural networks (SNNs) operate in an event-driven manner, where accumulate operations are triggered in postsynaptic neurons only when input spikes arrive.
Therefore, we further measure the average input firing rate of each SNN layer and combine it with the number of time steps to estimate the equivalent number of accumulate operations (ACs) in the SNN.
The number of computational operations of ANN and SNN ($\mathrm{O}_{\mathrm{ANN}}$ and $\mathrm{O}_{\mathrm{SNN}}$) can be described as:
\begin{equation}
\label{eq:35}
\mathrm{O}_{\mathrm{ANN}}=\sum_{l=1}^{L}O_{l},
\mathrm{O}_{\mathrm{SNN}} =\mathrm{T}\sum_{l=1}^{L}O_{l}r_{l,\mathrm{in}}
\end{equation}
where $O_{l}$ denotes the number of computational operations in the $l$-th layer, $r_{l,\mathrm{in}}$ denotes the average input firing rate of the $l$-th SNN layer and $\mathrm{T}$ denotes the time step of the SNN.
Then, the per inference of the ANN and SNN ($\mathrm{E(ANN)},\mathrm{E(SNN)}$) is estimated as:
\begin{equation}
\label{eq:36}
\mathrm{E(ANN)}=\mathrm{O(ANN)}\mathrm{E_{MAC}},
\mathrm{E(SNN}) =\mathrm{O(SNN)}\mathrm{E_{AC}}
\end{equation}
where $E_{\mathrm{MAC}}$ and $E_{\mathrm{AC}}$ denote the energy costs of a single MAC operation and a single AC operation, respectively.

For networks deployed on real hardware platforms, the per inference energy consumption is obtained by dividing the measured dynamic power by the inference speed.
The dynamic power was defined as the difference between the operating power and the idle power of each hardware platform during inference. 
Specifically, the power consumption of the NVIDIA GeForce RTX GPU was recorded using the NVIDIA System Management Interface (nvidia-smi).
The power consumption of the NVIDIA Jetson Orin NX was measured through its onboard power monitoring interface,  whereas that of the Intel Kapoho Bay neuromorphic processor was measured using an external power meter. 
To measure inference speed, each network was repeatedly evaluated on the corresponding hardware platform, and the average inference speed was reported.

\paragraph*{Hardware deployment of the proposed system}
The entire system is implemented on \textsc{Ubuntu 20.04}, with the Robot Operating System (\textsc{ROS}) used for information transfer and communication.
We use the official Prophesee ROS driver~\cite{prophesee_ros} and \textsc{OpenEB} 4.6.2~\cite{openeb}
for event stream acquisition.
To facilitate real-time event stream processing, the original driver was modified to directly output event frames, and the resolution was reduced from 1280 $\times$ 720 to 320 $\times$ 180.
The compilation and deployment of the navigation spiking network, as well as communication between the host computer and the neuromorphic processor, are accomplished using the \textsc{NxSDK} 1.0.0 library provided by Intel’s Neuromorphic Computing Research Team. 
The total network is trained using \textsc{PyTorch} on an NVIDIA RTX 4090 GPU.
Before real-world flight, a portion of real-world data is collected for depth fine-tuning.
Following the data acquisition method proposed by Anish Bhattacharya et al.~\cite{bhattacharya2024monocular}, we collected real-world event streams and the corresponding depth ground truth for fine-tuning.
Further details on data collection and processing are provided in the supplementary material.
We use a downward-facing Intel RealSense T265 to provide pose estimation for the system.
For flight control, we adopt the open-source controller~\cite{fast_drone} developed by the FAST Lab at Zhejiang University.





\clearpage 

%
\bibliography{science_template} 
\bibliographystyle{sciencemag}

%
%
%
%
%
%


\section*{Acknowledgments}
\paragraph*{Funding:}
This work was supported in part by the National Natural Science Foundation of China under Grants 62073066,  in part by Intel Neuromorphic Research Community (INRC) Grant Award (RV2.137.Fang), 
and in part by the Liaoning Provincial Future Industry Research and Development Program under Grant 2026JH40/10200065.
\paragraph*{Author contributions:}
J.J. and Z.F. proposed the architecture of the integrated neuromorphic visual navigation system. 
J.J. implemented the algorithms in both simulation and real-world environments, and adapted the spiking navigation network to the Intel Loihi processor. 
A.Z. designed the complete hardware system. J.J. and A.Z. integrated the neuromorphic system onto the MAV and conducted the real-world experiments. All authors contributed to the analysis and interpretation of the results. 
J.J. wrote the manuscript with input from all authors. 
Z.F. conceived the concept of the integrated neuromorphic navigation system and supervised the entire project.
\paragraph*{Competing interests:}
There are no competing interests to declare.
\paragraph*{Data and materials availability:}
The source code, trained model weights, and datasets supporting the findings of this study are currently being curated for public release. They will be deposited in a public and persistent repository upon acceptance and made publicly available no later than the publication of the article. Repository links and persistent identifiers will be provided in the final version of the manuscript.


\subsection*{Supplementary materials}
Materials and Methods\\
Figures. S1 to S4\\
Tables S1 to S3\\
References \textit{(55-\arabic{enumiv})}\\ 
Movie S1 to S2



\newpage


\renewcommand{\thefigure}{S\arabic{figure}}
\renewcommand{\thetable}{S\arabic{table}}
\renewcommand{\theequation}{S\arabic{equation}}
\renewcommand{\thepage}{S\arabic{page}}
\setcounter{figure}{0}
\setcounter{table}{0}
\setcounter{equation}{0}
\setcounter{page}{1} 


\begin{center}
\section*{Supplementary Materials for\\ \scititle}

Junjie~Jiang,
Ankang~Zhou,
Zheng~Fang$^{\ast}$\\
\small$^\ast$Corresponding author. Email: fangzheng@mail.neu.edu.cn\\
\end{center}

\subsubsection*{This PDF file includes:}
Materials and Methods\\
Figures S1 to S4\\
Tables S1 to S3\\
References \textit{(55-\arabic{enumiv})}\\
Captions for Movies S1 to S2%

\subsubsection*{Other Supplementary Materials for this manuscript:}
Movies S1 to S2

\newpage


\subsection*{Materials and Methods}
\begin{figure} 
	\centering
	\includegraphics[width=\textwidth]{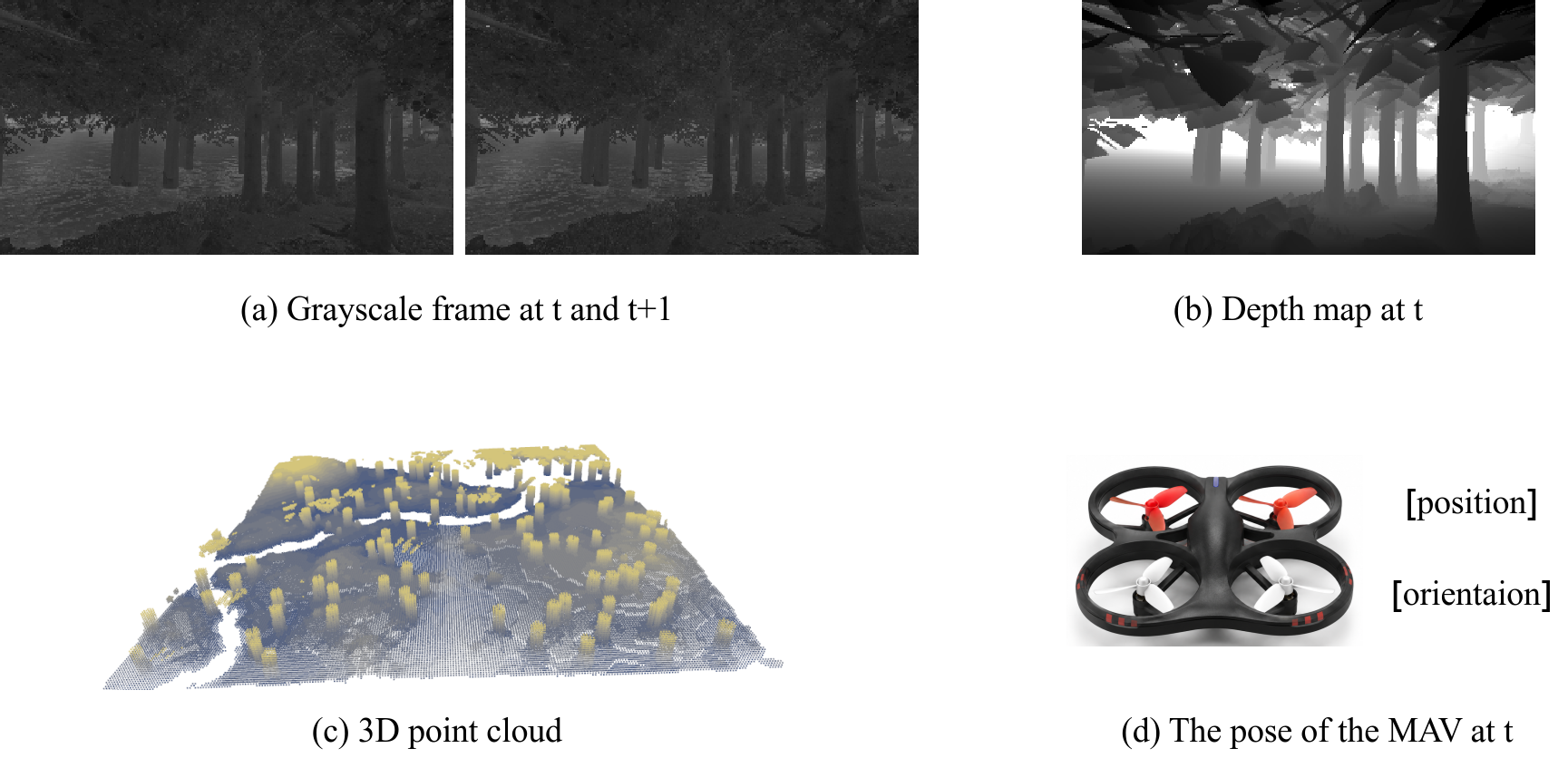} 

	\caption{\textbf{Composition of the training dataset.} Each training dataset includes a complete point cloud map of the environment, along with continuous grayscale images, corresponding depth maps, and MAV poses collected from multiple locations.}
	\label{fig:s1} 
\end{figure}

\paragraph*{Training data collection}
We use the \textsc{Flightmare} simulator~\cite{song2020flightmare} to collect training data in forest-like environments. 
For each scene, we collect four types of data: a pair of consecutive grayscale images, the corresponding depth image, the MAV pose, and the 3D point cloud of the overall environment, as shown in Fig.~\ref{fig:s1}. 
The consecutive grayscale image pair is generated by applying a small perturbation to the MAV pose, enabling subsequent event stream rendering with \textsc{vid2e}~\cite{gehrig2020video}. 
In total, we generate 20 different training scenarios and collect 2,500 grayscale image pairs for each scenario, along with their corresponding depth images and MAV poses.

\begin{figure} 
	\centering
	\includegraphics[width=0.8\textwidth]{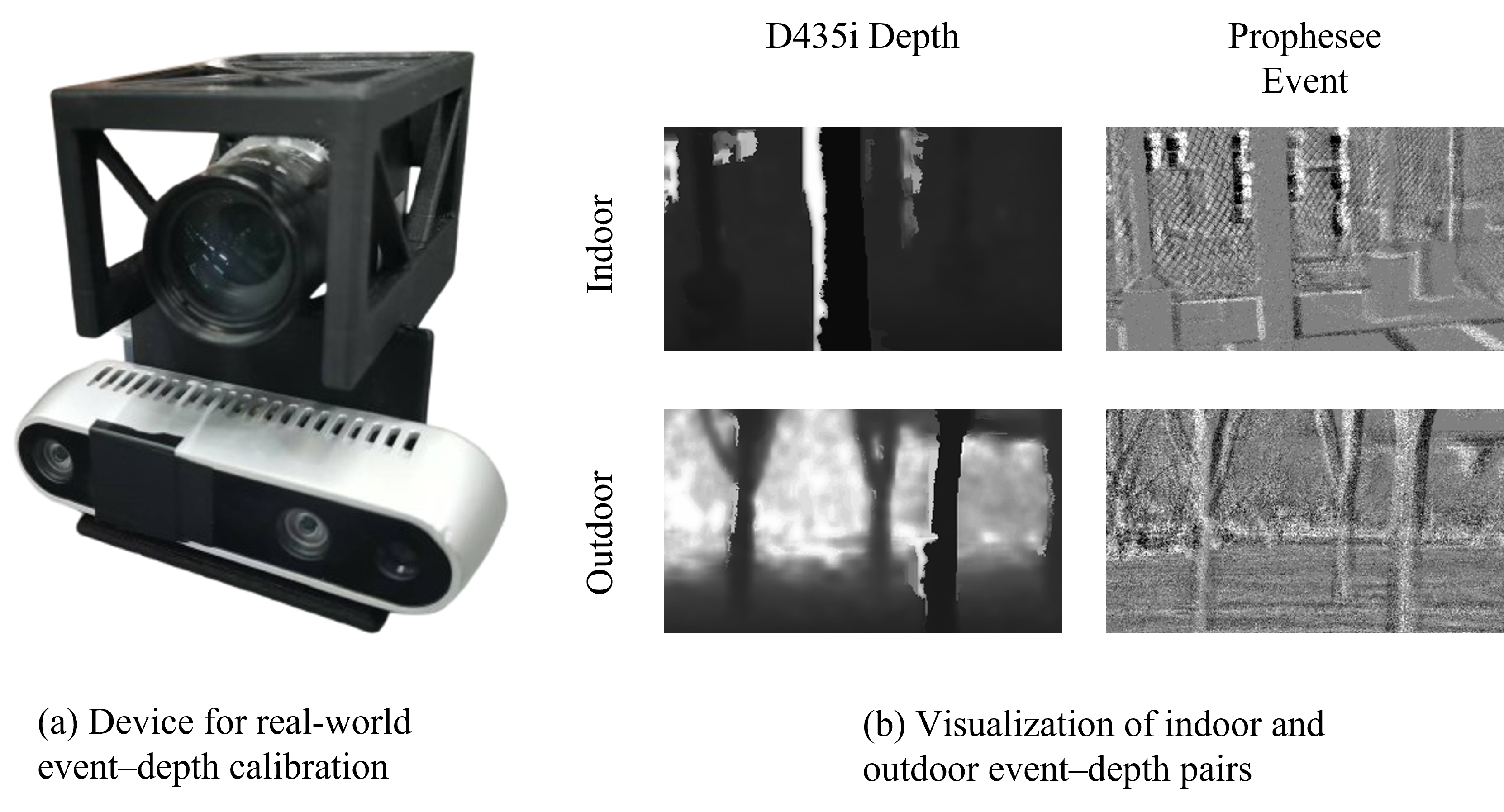} 

	\caption{\textbf{Event–depth calibration for fine-tuning in the real world.} The Prophesee EVK4 and Intel RealSense D435i are rigidly mounted together to enable joint camera calibration. The figure also shows representative event frames and their corresponding depth maps acquired in indoor and outdoor environments for real-world fine-tuning of the spiking depth estimation network.}
	\label{fig:s2} %
\end{figure}

\paragraph*{Real-data fine-tuning of the spiking depth estimation network}
To transfer the model from simulation to real-world scenarios, depth supervision corresponding to real event frames is required.
However, the Prophesee EVK4 event camera does not provide depth measurements.
Therefore, an Intel RealSense D435i is used to acquire depth ground truth corresponding to the event data, as shown in Fig.~\ref{fig:s2}. 
During data collection, depth images and event streams are synchronously recorded using the official Intel RealSense ROS driver~\cite{realsense_ros} and a third-party Metavision ROS driver~\cite{metavision_ros}, respectively.
Since the D435i and EVK4 differ in resolution and field of view, we first perform joint calibration of the dual-camera system using \textsc{E2Calib}~\cite{muglikar2021calibrate} and \textsc{Kalibr}~\cite{maye2013self} to obtain the camera intrinsic parameters, extrinsic parameters, and distortion coefficients. 
Then, event streams and depth images are collected in both indoor and outdoor real-world scenes. 
The continuous event streams are accumulated into event frames corresponding to the depth images according to ROS timestamps. 
Based on the calibration parameters, the event intensity frames and depth images are further geometrically rectified and spatially aligned to establish pixel-level correspondence between them. 
Finally, the aligned event frames and depth images are uniformly resized from 1280 × 720 to 320 × 180 for fine-tuning the depth estimation module in real-world scenarios.

\begin{table}[htbp]
\centering
\caption{\textbf{Parameter settings used in the experiments.} The table summarizes the principal hyperparameters used for network training, trajectory-loss computation, spiking depth estimation, spiking feature extraction, and the spiking navigation network.}
\renewcommand{\arraystretch}{1.15}
\begin{tabular}{c c c c}
\hline
\textbf{Types} & \textbf{Parameters} & \multicolumn{1}{c}{\textbf{Descriptions}} & \textbf{Values} \\
\hline

& Training epoch & -- & 100 \\
\shortstack{Experimental\\setting}& Batch size & -- & 16 \\
& $r_0$ & Initial learning rate & $1.5\times10^{-4}$ \\
& $\mathrm{C}$ & Trajectory time constant & $8$ \\
\hline

& $w_{s}$ & The coefficient for smooth item& $4\times10^{-4}$ \\
\shortstack{Loss\\function} & $w_{o}$ & The coefficient for collision item& $10^{-3}$ \\
& $w_{g}$ & The coefficient for goal item & $2\times10^{-4}$ \\
\hline

& $\mathrm{T}$ & Time step & 4 \\
\shortstack{Spiking depth\\estimation network} & $\tau$ & Membrane potential time constant & 2 \\
& $v_{\mathrm{th}}$ & spike-triggered threshold & 0.5 \\
\hline

& $\mathrm{T}$ & Time step & 4 \\
\shortstack{Spiking feature\\extraction network} & $\tau$ & Membrane potential time constant & 2 \\
& $v_{\mathrm{th}}$ & spike-triggered threshold & 0.4 \\
\hline
& $N_{\text{in}}$ & The number of gassusian distributions & 2 \\
& $N_{\text{out}}$ & The number of decoding neurons & 10 \\
&$u_{\mathrm{th, en}}$ & Encoding spike-triggered threshold & 0.999 \\
\shortstack{Spiking navigation\\network}&$\mathrm{T}$ & Time step & 5 \\
&$\delta_{\mathrm{curr}}$ & membrane current decay coefficient & 0.5 \\
& $\delta_{\mathrm{volt}}$ & membrane voltage decay coefficient & 0.75 \\
& $v_{\mathrm{th}}$ & spike-triggered threshold & 0.5\\
\hline
\end{tabular}
\label{Table:s1}
\end{table}


\begin{table}[t]
\centering
\caption{\textbf{Layer-wise computational statistics of the spiking U-Net.} For each layer, the table reports the corresponding ANN multiply–accumulate operations (MACs), the average input firing rate measured in the SNN, and the resulting number of spike-triggered accumulate operations (ACs). The AC count is obtained by weighting the corresponding layer operations by the average input firing rate and the number of SNN time steps.}
\renewcommand{\arraystretch}{0.85}
\setlength{\tabcolsep}{4pt}
\begin{tabular}{ccccc}
\hline
\textbf{Module} & \textbf{Operations} & \textbf{Corresponding MACs} & \textbf{Average input firing rate} & \textbf{ACs} \\
\hline

&Conv-1 & 4.72M & 0.3696 &6.98M \\
\raisebox{2.5ex}{Encode-1}&Conv-2 &37.75M &0.2234 &33.73M \\
\hline

&Conv-1 &18.87M &0.3711 &28.01M \\
\raisebox{2.5ex}{Encode-2}&Conv-2  &37.75M &0.1968 &29.72M \\
\hline

&Conv-1  &18.87M &0.3613 &27.27M \\
\raisebox{2.5ex}{Encode-3}&Conv-2  &37.75M &0.2102 &31.74M \\
\hline

&Conv-1   &18.87M &0.3729 &28.15M \\
\raisebox{2.5ex}{Encode-4}&Conv-2  &37.75M &0.1992 &30.08M \\
\hline

&Conv-1  &9.44M &0.3899 &14.72M \\
\raisebox{2.5ex}{Bottleneck}&Conv-2  &9.44M &0.1896 &7.16M \\
\hline

UpBlock-4 &DeConv  &4.19M &0.1925 &3.23M \\
\hline

&Conv-1  &37.75M &0.1802 &27.21M \\
\raisebox{2.5ex}{Decode-4}&Conv-2  &9.44M &0.1896 &7.16M \\
\hline

UpBlock-3 &DeConv  &4.19M &0.2091 &3.50M \\
\hline

&Conv-1  &37.75M &0.1840 &27.78M \\
\raisebox{2.5ex}{Decode-3}&Conv-2  &9.44M &0.2135 &8.06M \\
\hline

UpBlock-2 &DeConv  &4.19M &0.2285 &3.83M \\
\hline

&Conv-1  &37.75M &0.1915 &28.92M \\
\raisebox{2.5ex}{Decode-2}&Conv-2  &9.44M &0.2318 &8.75M \\
\hline

UpBlock-1 &DeConv  &4.19M &0.2757 &4.62M \\
\hline

&Conv-1  &75.50M &0.2117 &63.93M \\
\raisebox{2.5ex}{Decode-1}&Conv-2  &37.75M &0.2087 &31.51M \\
\hline

Depth Head &Conv  &0.26M &0.2826 &0.29M \\
\hline
\end{tabular}
\label{Table:s2}
\end{table}

\begin{table}[t]
\centering
\caption{\textbf{Layer-wise computational statistics of the spiking feature extraction network.} For each layer, the table reports the corresponding ANN multiply–accumulate operations (MACs), the average input firing rate of the SNN, and the resulting number of spike-triggered accumulate operations (ACs). The AC count is calculated based on the corresponding operation count, average input firing rate, and number of SNN time steps.}
\label{tab:snn_unet_computation}
\renewcommand{\arraystretch}{1.15}
\setlength{\tabcolsep}{6pt}
\begin{tabular}{ccccc}
\hline
\textbf{Module} & \textbf{Operations} & \textbf{Corresponding MACs} & \textbf{\shortstack{Average input\\firing rate}} & \textbf{ACs} \\
\hline

Stem &Conv &12.85M &1 &51.40M \\
\hline

&Conv-1 &9.44M &0.2870 &10.84M \\
ResBlock-1.1 &Conv-2 &9.44M &0.2672 &10.09M \\
&Conv(Downsample) &1.05M &0.2870 &1.21M \\
\hline
\raisebox{-2.0ex}{ResBlock-1.2} &Conv-1 &9.44M &0.4588 &17.32M \\
&Conv-2 &9.44M &0.2174 &8.21M \\
\hline

&Conv-1 &1.18M &0.5787 &2.73M \\
ResBlock-2.1 &Conv-2 &2.36M &0.2211 &2.09M \\
&Conv(Downsample) &0.13M &0.5787 &0.30M \\
\hline
\raisebox{-2.0ex}{ResBlock-2.2} &Conv-1 &2.36M &0.0810 &0.76M \\
&Conv-2 &2.36M &0.1781 &1.68M \\
\hline
Feature Head &FC  &0.09M &0.1054 &0.04M \\
\hline
\end{tabular}
\vspace{2mm}
\label{Table:s3}
\begin{minipage}{0.96\linewidth}
\footnotesize
\textbf{Note:} The SEW residual fusion introduces only element-wise operations and is therefore excluded from the MAC/AC statistics.
\end{minipage}
\end{table}

\begin{figure} 
	\centering
	\includegraphics[width=\textwidth]{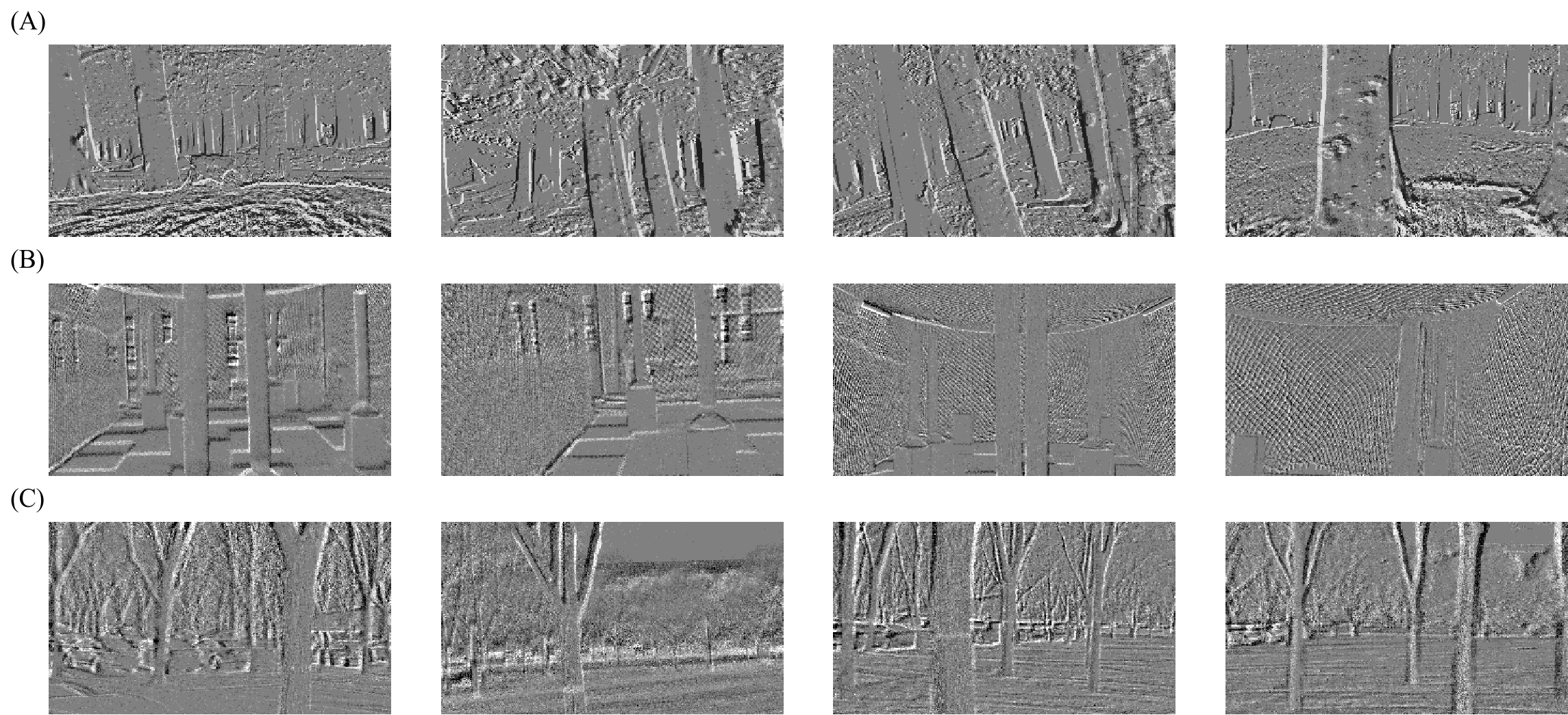} 

	\caption{\textbf{Event frames visualization across different environments.}
		(A) Simulated event frames. (B) Event frames captured in indoor environments. (C) Event frames captured in outdoor environments. For improved visualization, the event-frame values are amplified by a factor of 10 and then displayed as positive and negative deviations from a baseline intensity value of 127.}
	\label{fig:s3}
\end{figure}

\begin{figure} 
	\centering
	\includegraphics[width=\textwidth]{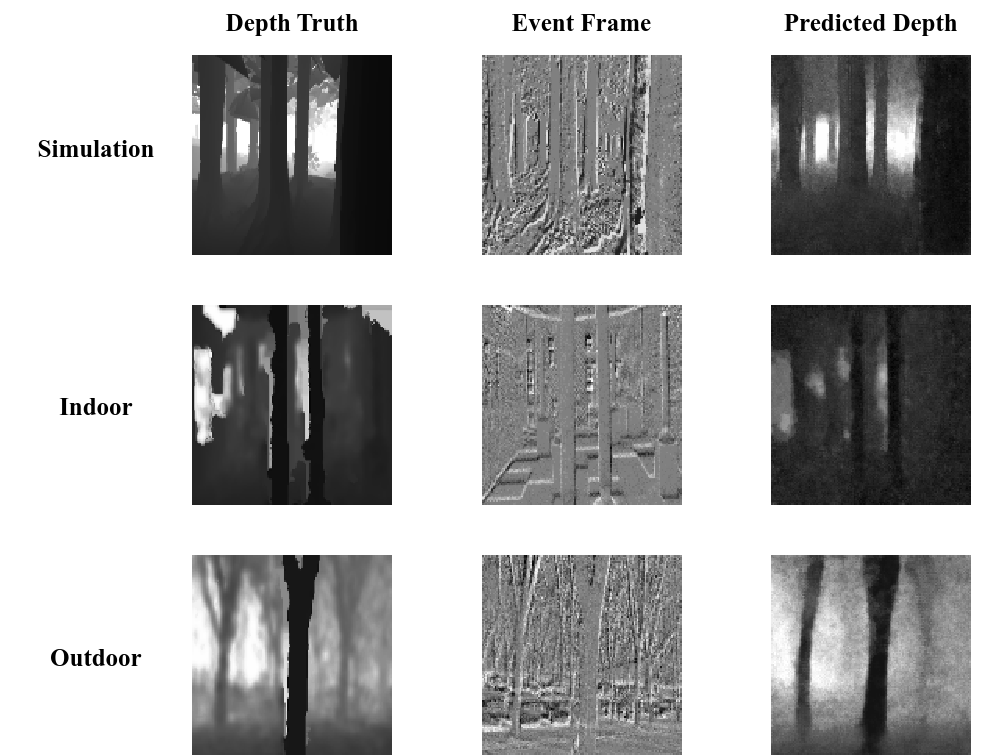} 

	\caption{\textbf{Visualization of predicted depth maps.}
	We present the depth ground truth, event frames, and predicted depth maps in simulation, indoor, and outdoor environments, respectively. For improved visualization, the event-frame values are amplified by a factor of 10 and then displayed as positive and negative deviations from a baseline intensity value of 127.}
	\label{fig:s4}
\end{figure}


\clearpage 

\paragraph{Caption for Movie S1.}
\textbf{Navigation performance in simulated environments.}
The left panel shows predicted depth, event frames, and ground-truth depth, while the right panel shows obstacle distributions and the corresponding MAV trajectories.
Both continuous navigation and navigation performance under different maximum flight speeds are shown.
The video demonstrates continuous navigation and navigation at different maximum flight speeds.

\paragraph{Caption for Movie S2.}
\textbf{Navigation performance in real-world environments.} Real-world MAV navigation in indoor and outdoor environments with diverse obstacle configurations. The left panel shows predicted depth and event frames, while the right panel presents the MAV’s real-world navigation performance.



\end{document}